\documentclass[10pt,twocolumn,letterpaper]{article}
\usepackage{authblk}

\usepackage[pagenumbers]{cvpr} %

\usepackage{graphicx}
\usepackage{amsmath}
\usepackage{amssymb}
\usepackage{booktabs}

\usepackage[pagebackref,breaklinks,colorlinks,citecolor=cyan]{hyperref}

\usepackage[capitalize]{cleveref}
\crefname{section}{Sec.}{Secs.}
\Crefname{section}{Section}{Sections}
\Crefname{table}{Table}{Tables}
\crefname{table}{Tab.}{Tabs.}

\usepackage{caption}
\usepackage{subcaption}

\usepackage{wrapfig}
\usepackage{xspace}
\usepackage{multirow}
\usepackage[export]{adjustbox}
\usepackage{makecell}

\newcommand{\point}{\mathbf{x}}
\newcommand{\direction}{\mathbf{d}}

\newcommand{\radiance}{\mathbf{c}}
\newcommand{\density}{\sigma}
\newcommand{\ray}{\mathbf{r}}
\newcommand{\latent}{\mathbf{l}}

\newcommand{\loss}{\mathcal{L}}
\newcommand{\bundle}{\mathcal{R}}

\newcommand{\real}{\mathbb{R}}
\newcommand{\realthree}{\mathbb{R}^3}

\newcommand{\Section}[1]{Section~\ref{sec:#1}}
\newcommand{\Figure}[1]{Figure~\ref{fig:#1}}
\newcommand{\Equation}[1]{Equation~\ref{equ:#1}}

\newcommand{\Circle}[1]{\textcircled{\small{#1}}}
\newcommand{\inquote}[1]{``#1"}

\newcommand{\SSDNerf}{SSDNeRF\xspace} %
\newcommand{\MaskRCNN}{Mask-RCNN\xspace} %

\makeatletter
\newcommand\suppRef[1]{%
  \@ifundefined{r@#1}{%
    the supplementary material%
  }{%
    \cref{#1}%
  }%
}
\makeatother

\begin{document}
\title{\SSDNerf: Semantic Soft Decomposition of Neural Radiance Fields}

\author[1,2]{Siddhant Ranade\thanks{Work done while author was an intern at Meta.}}%
\author[2]{Christoph Lassner}
\author[2]{Kai Li}
\author[2]{Christian Haene}
\author[2]{Shen-Chi Chen}
\author[2]{Jean-Charles Bazin}
\author[2]{Sofien Bouaziz}
\affil[1]{University of Utah --- \texttt{sidra@cs.utah.edu}}
\affil[2]{Meta --- \texttt{\{classner,kai1,chaene,schen119,jcbazin,sofienb\}@fb.com}}

\maketitle
\begin{abstract}

Neural Radiance Fields (NeRFs) encode the radiance in a scene parameterized by the scene's plenoptic function. %
This is achieved by using an MLP together with a mapping to a higher-dimensional space, and has been proven to capture scenes with a great level of detail. %
Naturally, the same parameterization can be used to encode additional properties of the scene, beyond just its radiance. %
A particularly interesting property in this regard is the semantic decomposition of the scene. %
We introduce a novel technique for semantic soft decomposition of neural radiance fields (named \SSDNerf) which jointly encodes semantic signals in combination with radiance signals of a scene. %
Our approach provides a \emph{soft} decomposition of the scene into semantic parts, enabling us to correctly encode multiple semantic classes blending along the same direction---an impossible feat for existing methods. %
Not only does this lead to a detailed, 3D semantic representation of the scene, but we also show that the regularizing effects of the MLP used for encoding help to improve the semantic representation. %
We show state-of-the-art segmentation and reconstruction results on a dataset of common objects and demonstrate how the proposed approach can be applied for high quality temporally consistent video editing and re-compositing on a dataset of casually captured selfie videos.\footnote{See \url{https://www.siddhantranade.com/research/2022/12/06/SSDNeRF-Semantic-Soft-Decomposition-of-Neural-Radiance-Fields.html} for the supp.\ video.} %

\end{abstract}
\section{Introduction}

Semantic analysis of scenes has been a subject of study since the early days of computer vision~\cite{SzeliskiBook}.
Traditionally, the problem has been posed as 2D image segmentation or matting in 2D image space. These methods have been shown to %
facilitate numerous image editing applications such as compositing foreground objects onto novel backgrounds.
However, with the increasing popularity of 3D neural scene representations and image based rendering, lifting semantics to the 3D world becomes increasingly important and opens a wide range of novel image/video editing applications yet to be invented. 

Early approaches dedicated to 3D semantic segmentation mainly focused on \inquote{traditional} 3D geometry representations like meshes, point clouds or voxel grids.
For image editing applications, such approaches would be insufficient, since the color of object boundaries in 2D images are often a combination of foreground and background colors.
In the 2D image domain, this is addressed using alpha matting where an alpha value per pixel defines how the foreground and background colors are mixed within a given pixel.
This idea can be extended into 3D soft semantic segmentation, where multiple classes can be present for a single pixel:
we need to take into account that a camera ray passes through several semantic classes in 3D space.
Some classes are visible in the final image, but some are blocked by geometry closer to the camera and others are partially visible.
All these interactions need to be taken into account correctly, leading to a final rendered result with smooth segmentation.

\begin{figure*}[t!]
    \centering
    \includegraphics[width=0.99\linewidth]{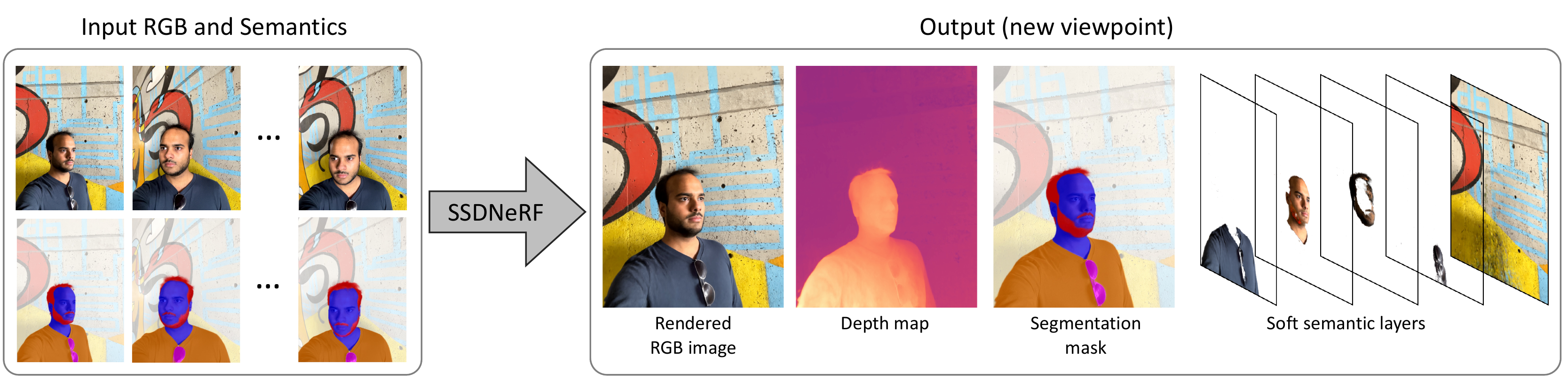}
    \caption{Given a set of RGB images and their segmentation masks computed by a neural network such as \MaskRCNN, \SSDNerf automatically renders an image and its depth map from a novel viewpoint, along with its soft segmentation and soft semantic decomposition. \SSDNerf decomposes a volumetric representation of the scene, thereby \textit{lifting} semantic segmentation from 2D to 3D.}
    \label{fig:overview}
\end{figure*}

This is similar in spirit to material density and radiance blending in neural radiance fields, %
which has been shown 
to be well handled
by a multi-layer perceptron (MLP).
We extend the formulation of classical neural radiance fields to incorporate such a \emph{soft blending} and model densities and colors for each class separately --- allowing for an accurate model of complex scenes with several semantic layers. 

In this paper, we introduce a \emph{semantic soft decomposition} of neural radiance fields, \SSDNerf.
As illustrated in \Figure{overview}, given a set of images and the results of a 2D semantic segmentation network, we learn a semantic soft decomposition of the 3D scene.
This semantic 3D lifting allows the rendering of each semantic layer from a novel view.
With our proposed method, we are able to not only trivially decompose scenes semantically, addressing classical computer vision problems such as foreground segmentation, but also edit and re-composite scenes using their 3D content.
Since the reasoning happens in 3D space, we can render our edited content consistently into 2D images resolving ambiguities that frame-by-frame 2D methods would be susceptible to.
To demonstrate these capabilities, we show various results and applications from casually captured videos and provide comparison with existing state-of-the-art methods.
Our typical capture scenario is a video selfie (see \Figure{overview}), and to handle involuntary motions of the captured person, we base our \SSDNerf implementation on deformable neural radiance fields frameworks~\cite{park2021nerfies}. %

Our contributions are: \Circle{1} a novel approach to decompose neural radiance fields into a set of \emph{soft} semantic layers; \Circle{2} a set of losses improving the geometric quality of the reconstructed layers; \Circle{3} a system to manipulate free viewpoint videos in a temporal consistent manner while respecting fine details and preserving view-dependent effects.

\section{Related Work}
Semantic scene analysis as well as view synthesis are both well-established fields. %
Our approach builds upon ideas from semantic segmentation and (compositional) neural radiance fields that we review in the following sections.

\subsection{Neural Radiance Fields}
Whereas the original neural radiance field~\cite{mildenhall2020nerf} implementation focuses on static scenes, we aim to reconstruct scenes captured in a casual setting with a single smartphone camera and where the subject might not be entirely static.
Non-rigid NeRF~\cite{tretschk2021nonrigid} allows for scenes with limited user motion, but makes few assumptions about the motion field and introduces a `rigidity' score for the material.
Nerfies~\cite{park2021nerfies} has been specifically designed to capture selfie videos into deformable NeRF and uses a restricted deformation model to support and account for minimal, involuntary user motion.
This opens up additional editing capabilities (see~\cite{park2021nerfies}), hence we base our deformation model on this work.
Many other deformable NeRF approaches have been developed based on requirements we do not aim to request from users such as multi-view data~\cite{dnerf} or specific sensors~\cite{torf}.
Another line of work uses explicit face models~\cite{nerface,adnerf} or body models~\cite{narf,animatablenerf,neuralbody} for better regularization, and thus are applicable only for these specific scenarios. %
Our approach can be used for full body capture as well as general dynamic scenes and we do not rely on object specific priors.
A large number of NeRF extensions have been recently published, see \cite{dellaert2020neural} for a review. For example, Mip-NeRF 360~\cite{barron2021mipnerf360} and NeRF++~\cite{nerf++} extend NeRF to render un-bounded scenes, or scenes with a large difference in distance between the closest and farthest scene elements.
While it is trivial to integrate these techniques in our formulation, we did not find them essential to produce high quality results in our current setting. 
Some recent works aim to speed up training and inference time~\cite{sun2021direct,mueller2022instant,yu_and_fridovichkeil2021plenoxels}, and we base our method on~\cite{mueller2022instant}, demonstrating fast convergence to reconstruct per-scene radiance fields.

Multiple compositional NeRF approaches have been recently proposed to improve the scalability and efficiency of the rendering process \cite{rebain2020derf,xiangli2021citynerf,Lombardi21,wang2021head}. Closer to our work, a set of compositional approaches \cite{niemeyer2021giraffe,stelzner2021decomposing,yu2021unsupervised,elich2020semsup3dobjs,guo2020object} enable 3D scene decomposition and manipulation. However, none of these approaches are able to provide accurate soft semantic layers critical for high quality video editing.

\subsection{Semantic Segmentation}

Natural images can be described as the composition of multiple objects where each pixel belongs to one or more classes. Image matting approaches~\cite{Levin08matting,Levin08spectralmatting} try to revert this process by decomposing an image into soft layers and disentangling the blending of background and foreground resulting in a pixel color. These approaches are heavily used in image editing software and enable the automation of complex editing tasks such as background replacement. 

With the recent advances in deep learning, semantic segmentation techniques have improved significantly allowing to jointly segment a large number of semantic classes \cite{chen2017deeplab,he2017mask}. One drawback of these approaches is the generation of hard edges that cannot be easily used for image editing purpose. To overcome this issue, the semantic soft segmentation approach~\cite{Aksoy2018sss} uses a DeepLab feature extractor \cite{chen2017deeplab} to condition a spectral matting optimization allowing to decompose an image into soft semantic layers. The other related approach~\cite{Sbai20unsuperdecomp} decomposes images into simple layers defined by their color and a vector transparency mask.
While these technique produce detailed layers, they are designed for single 2D images and thus are not temporally stable, leading to flickering artifacts when applied to videos. In contrast, our approach uses a \MaskRCNN instance segmentation model \cite{he2017mask} as supervision but consolidates the per-image labels using the NeRF framework allowing us to generate soft labels that are spatiotemporally consistent.

Semantic segmentation has also been considered in 3D by semantically segmenting meshes \cite{valentin2015semanticpaint}, point-clouds \cite{hackel2016fast} or voxels \cite{sengupta2013urban}. By jointly optimizing for 3D geometry and semantic segmentation, the interdependence between these tasks can be utilized \cite{hane2016dense}. Describing data terms as potentials over viewing rays \cite{savinov2016semantic} can further enhance the accuracy of the obtained 3D reconstructions. End-to-end learning for semantic 3D reconstruction has also been considered \cite{tulsiani2017multi,hou20193d}. These methods focus on 3D geometry reconstruction and its semantic segmentation without taking into account novel view synthesis and image based rendering. In contrast, our approach incorporates 3D reconstruction, novel view synthesis and semantic segmentation all within the same network.

Semantic-NeRF \cite{Zhi21semanticnerf} is the closest approach to our work. They treat the semantic logits as additional radiance channels (identical to color). In contrast, our proposed approach explicitly models multiple classes of objects in the scene by re-deriving the volume rendering formulation for multiple classes, thereby automatically disentangling the representations of different classes. This enables more effective regularization of the opacity, and allows the rendering and manipulation of individual classes of objects.

\section{Method}
We propose a novel approach allowing the creation of neural radiance fields composed of soft semantic layers.
Instead of generating a single density and radiance field for the scene, our approach produces a density and radiance fields \emph{per semantic layer} to represent soft transitions between semantic classes (\Section{decomposition}).
We also introduce a set of regularizations taking advantage of this decomposition to generate high quality semantic layers  (\Section{losses}).

\subsection{Layer Decomposition}
\label{sec:decomposition}

\def\figScale{0.185}
\setlength\tabcolsep{0.005\textwidth}
\begin{figure*}[htb]%
\centering%
\begin{tabular}{cccccc}
\adjincludegraphics[width=\figScale\textwidth,trim={0 0 0 {0.25\height}},clip]{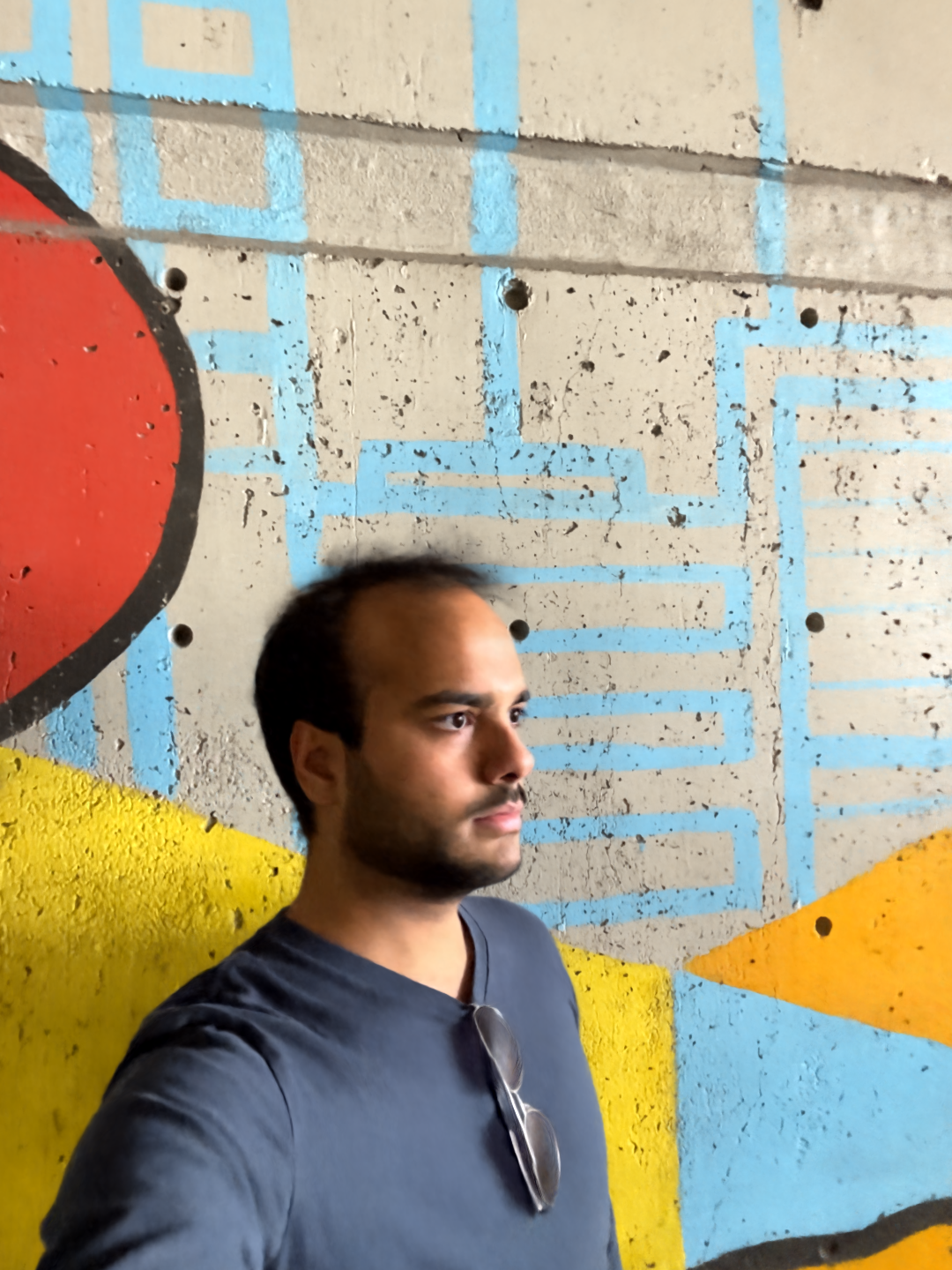}
&\adjincludegraphics[width=\figScale\textwidth,trim={0 0 0 {0.25\height}},clip]{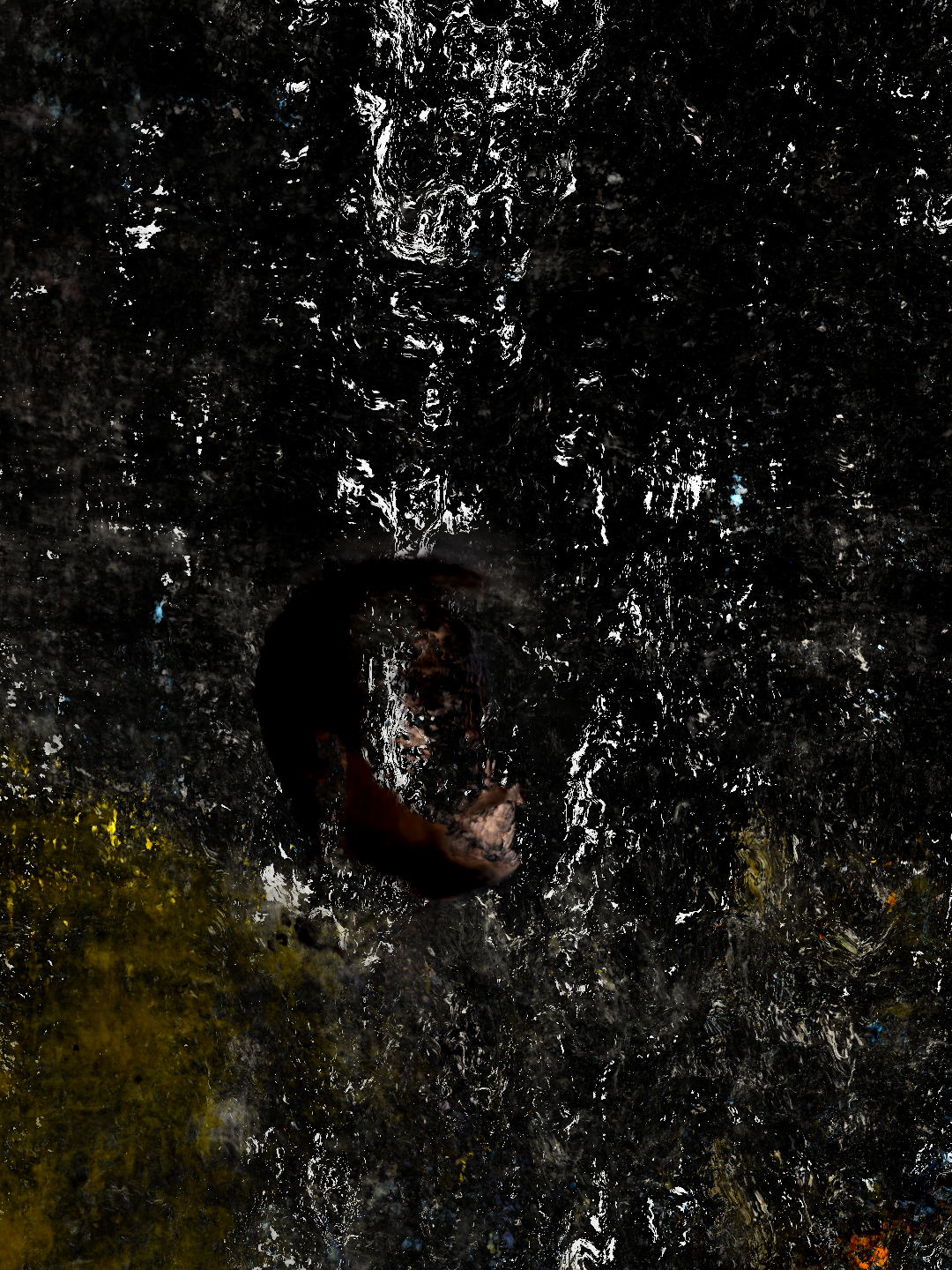}
&\adjincludegraphics[width=\figScale\textwidth,trim={0 0 0 {0.25\height}},clip]{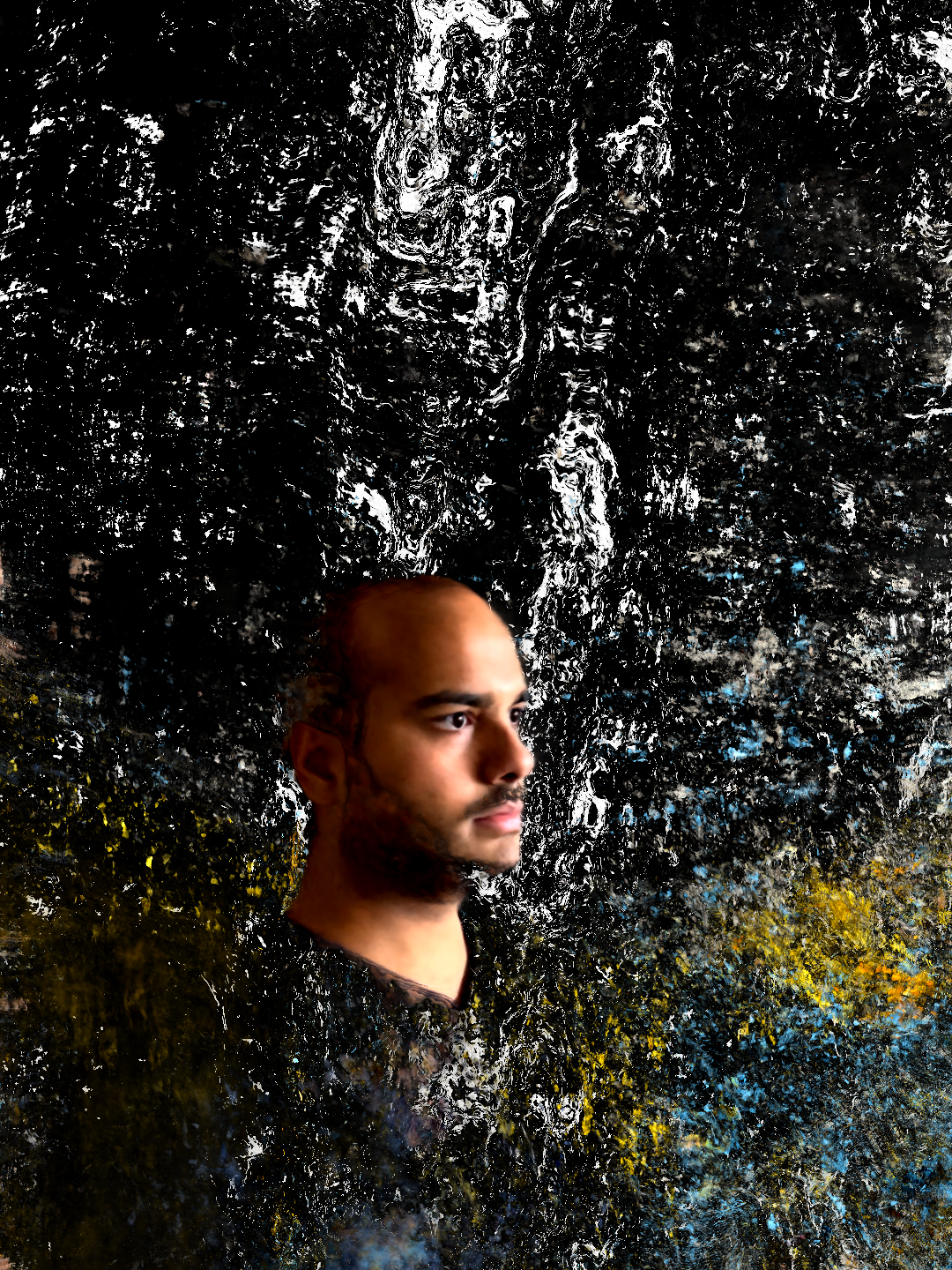}
&\adjincludegraphics[width=\figScale\textwidth,trim={0 0 0 {0.25\height}},clip]{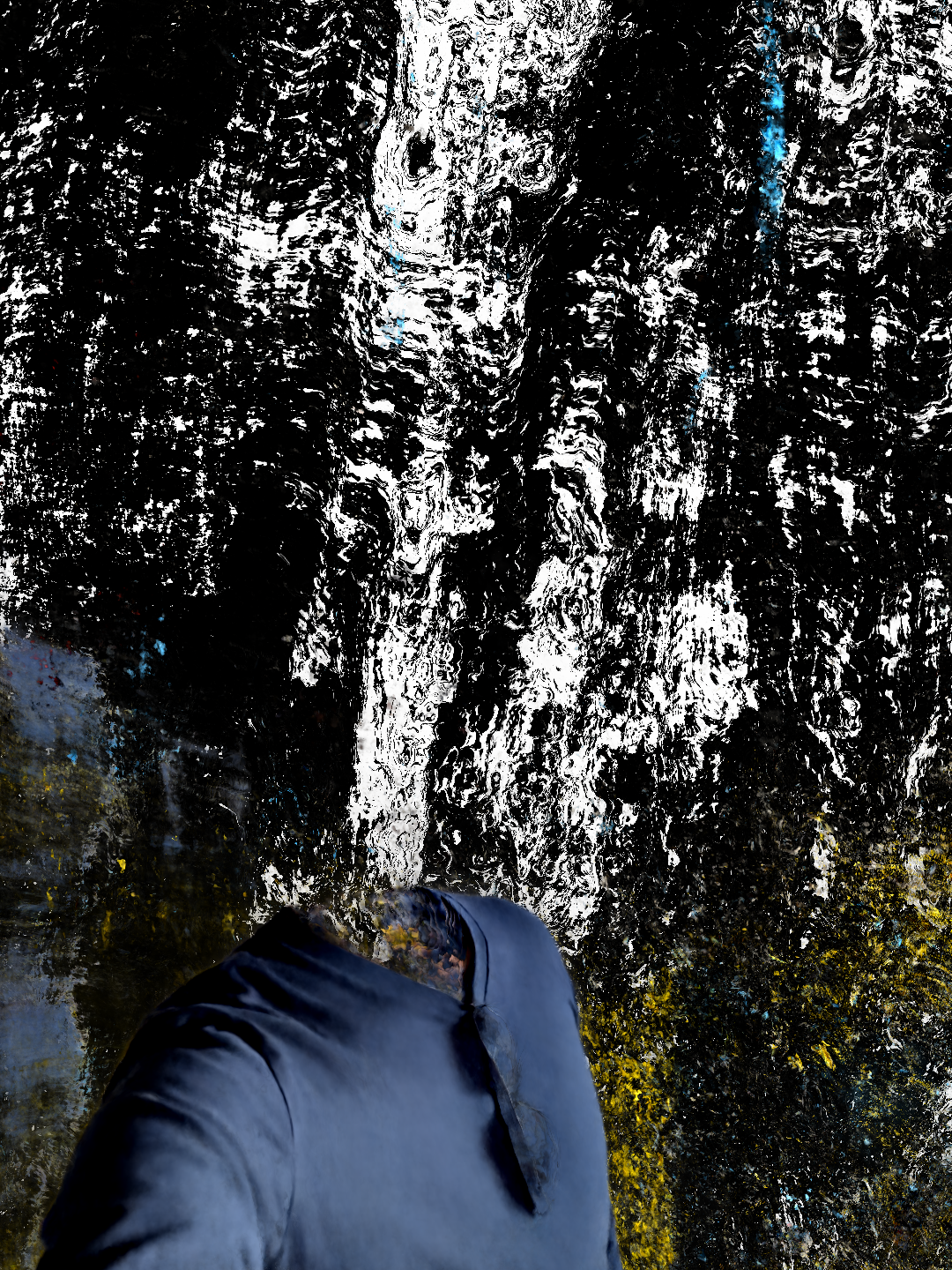}
&\adjincludegraphics[width=\figScale\textwidth,trim={0 0 0 {0.25\height}},clip]{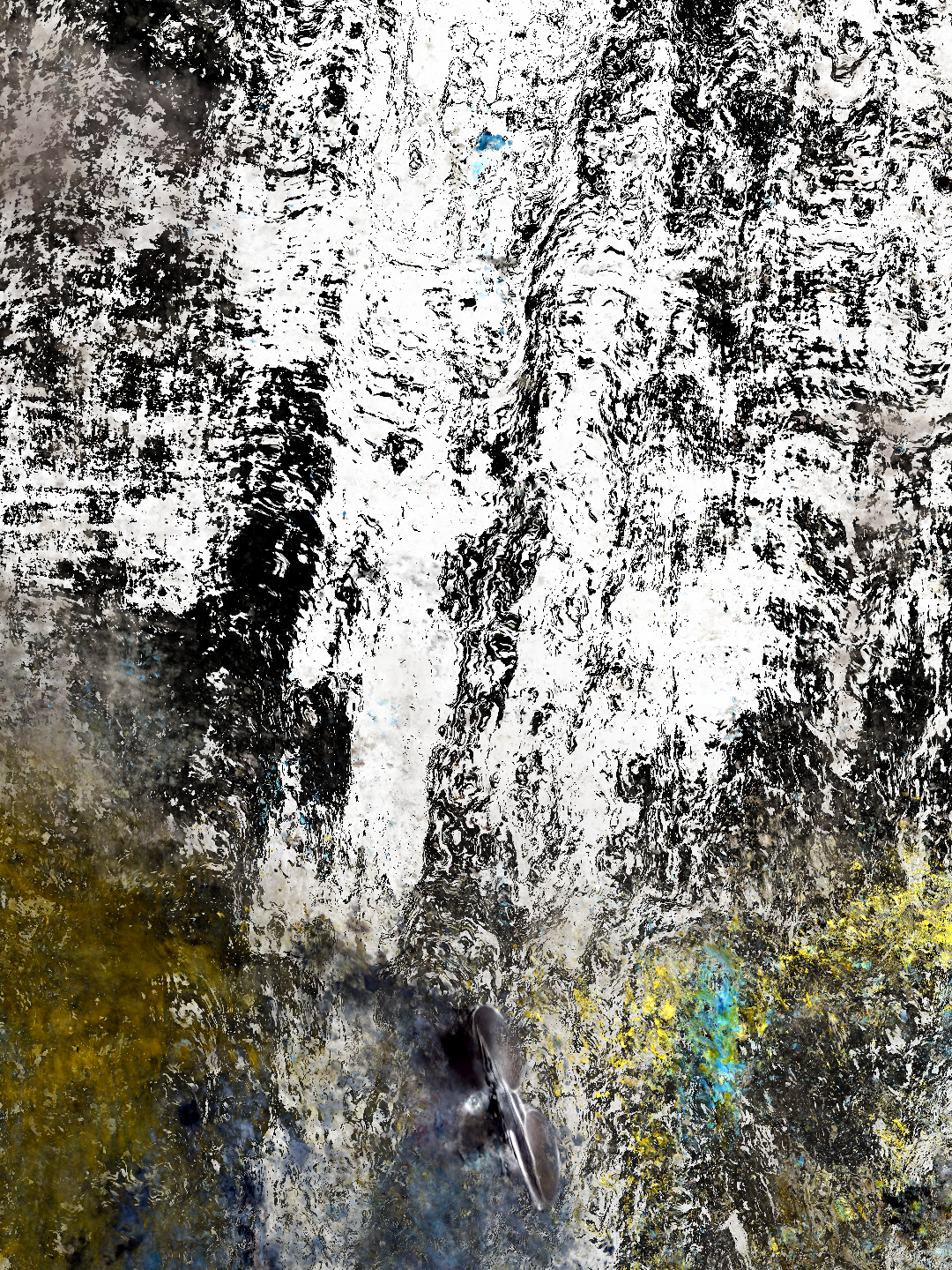} \\
\adjincludegraphics[width=\figScale\textwidth,trim={0 0 0 {0.25\height}},clip]{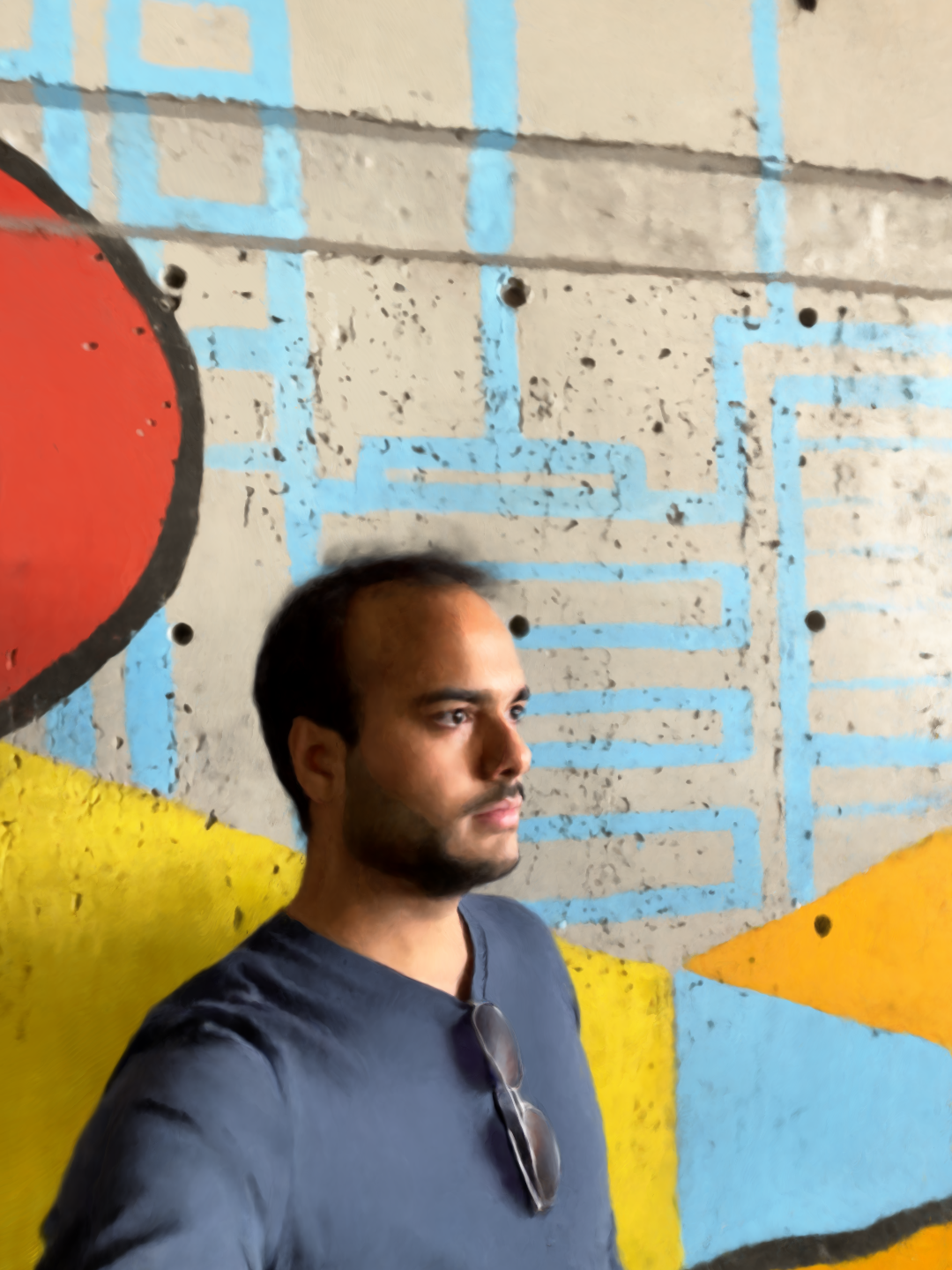}
&\adjincludegraphics[width=\figScale\textwidth,trim={0 0 0 {0.25\height}},clip,frame]{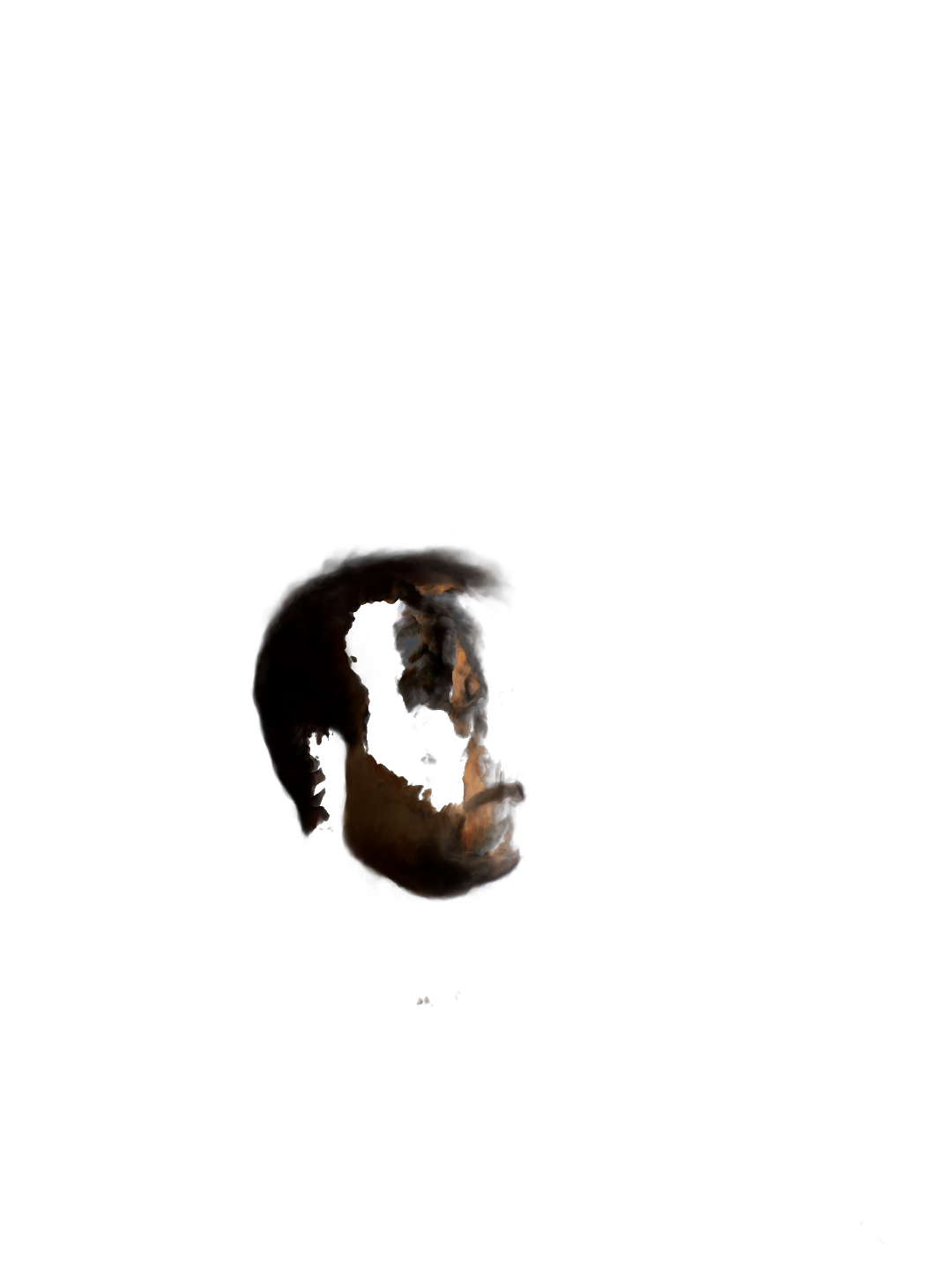}
&\adjincludegraphics[width=\figScale\textwidth,trim={0 0 0 {0.25\height}},clip,frame]{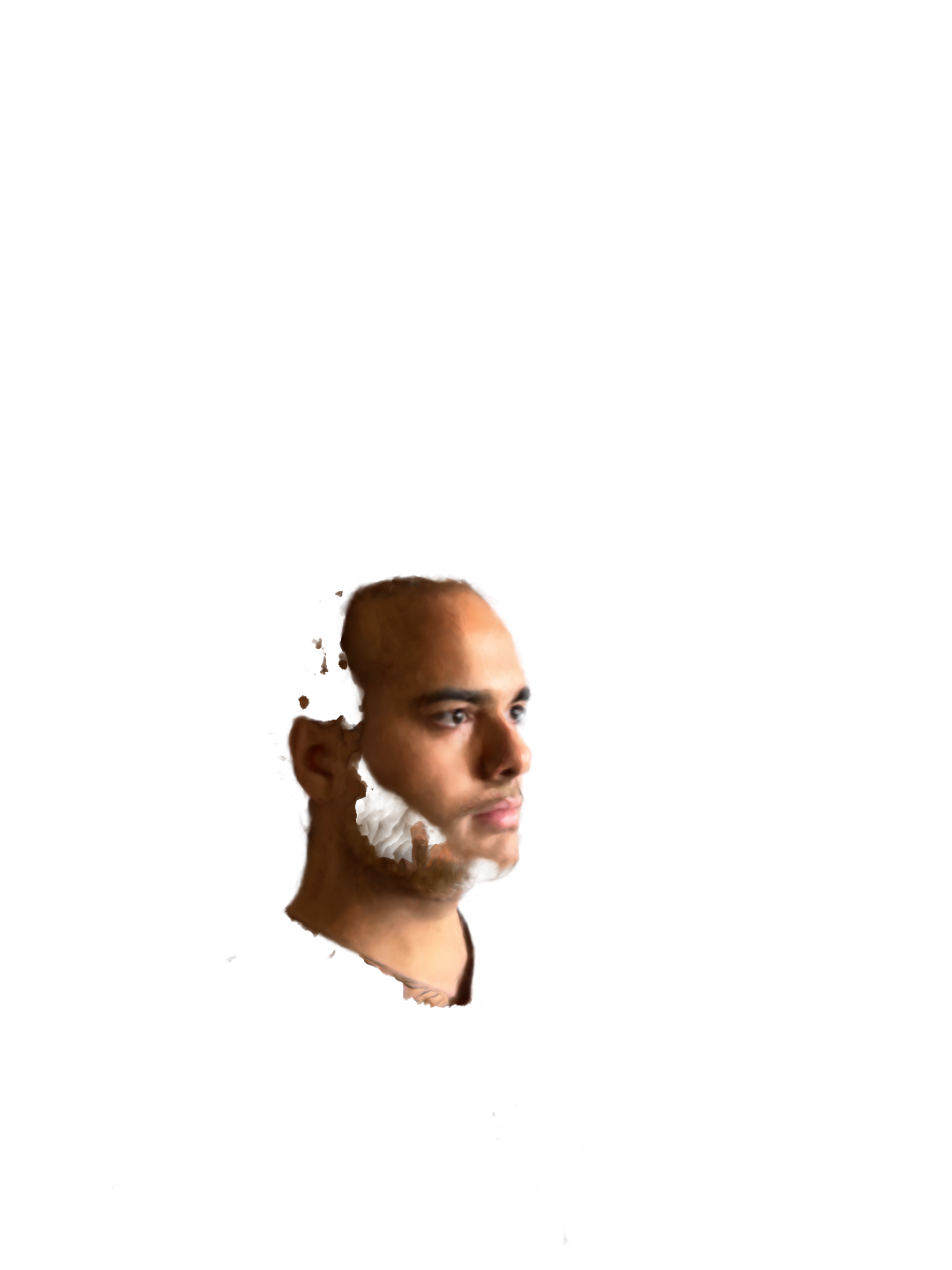}
&\adjincludegraphics[width=\figScale\textwidth,trim={0 0 0 {0.25\height}},clip,frame]{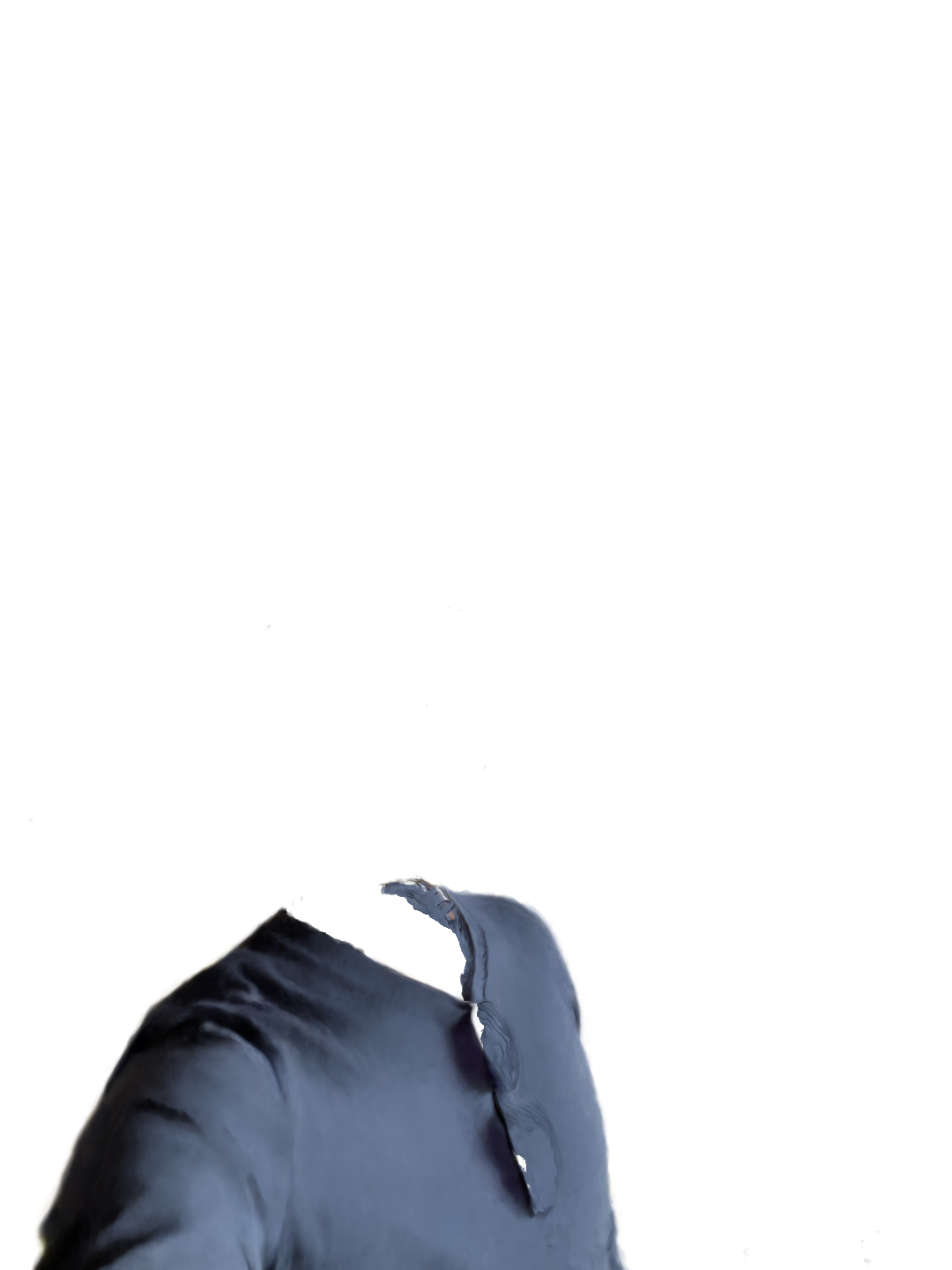}
&\adjincludegraphics[width=\figScale\textwidth,trim={0 0 0 {0.25\height}},clip,frame]{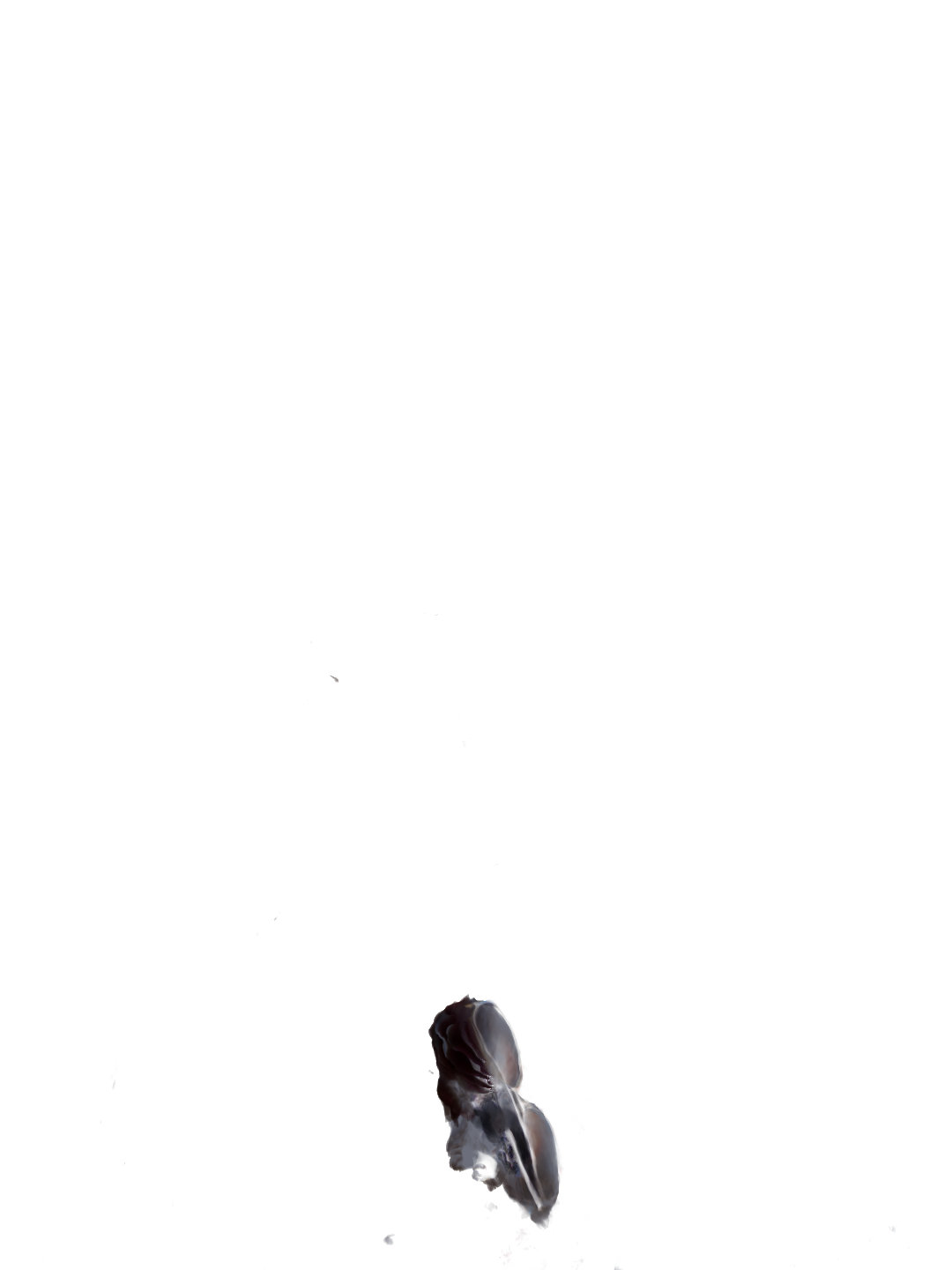} \\
RGB & Hair & Skin & T-shirt & Glasses \\
\end{tabular}
\vspace{-4mm}
\caption{Comparison of SNeRF~\cite{Zhi21semanticnerf} (top) and our \SSDNerf (bottom): rendered RGB image (leftmost column) and decomposed semantic layers (other columns). Contrary to our approach which is well regularized using sparsity priors, SNeRF predicts non-zero logits for all classes at all points including free space, leading to noisy layers.}
\label{fig:compare_snerf_layers}
\vspace{-3mm}
\end{figure*}

A neural radiance field (NeRF)~\cite{mildenhall2020nerf} is a continuous representation $F: (\point, \direction) \rightarrow (\radiance, \density)$, which maps a point $\point \in \realthree$ on a viewing ray with direction $\direction \in \realthree$ to an RGB color $\radiance \in \realthree$ and a density $\density \in \real$.
The function $F$ is parametrized by an MLP (multi-layer perceptron) and can be queried at arbitrary points allowing to render a volume by accumulating the color values along camera rays by using the density for alpha compositing. 

To decompose a NeRF into a set of $M$ semantic layers, our approach extends this formulation and generates a color and a density value \emph{per semantic layer}, i.e., $F : (\point, \direction) \rightarrow (\radiance^1, \density^1, \hdots, \radiance^{M}, \density^{M})$.
To disambiguate indices, we denote the index for the different layer $i$ in superscript, the index for the ray sample index $j, k$ in subscript notation---there are no powers required in the following formulas.
First, let's recall the compositing equation to accumulate $N$ samples along a ray~$\ray$ in the original NeRF formulation~\cite{mildenhall2020nerf}:
\begin{align} \label{equ:nerf_composition}
C(\ray) &= \sum_{j=1}^N T_j \left(1 - \exp\left(-\density_j\delta_j\right)  \right)\radiance_j, \\
&\text{where} \quad T_j = \exp\left(-\sum_{k=1}^{j-1}\density_k\delta_{k}\right), \nonumber
\end{align}
$\density_j$ and $\radiance_j$ are respectively the densities and colors of the $j^\text{th}$ sample point along ray $\ray$, and $\delta_k$ is the distance between the $(k-1)^\text{th}$ and $k^\text{th}$ ray samples.
In our work, this formulation remains unchanged, however, we aim to create a $\sigma$-weighted average across the different layers: hence, at the same point in space the resulting accumulated $\sigma$ value should be the sum of the material density of all the layers and the resulting color should be weighted by the respective layer densities.
The density $\sigma$ is exactly the aforementioned sum, and the color $\radiance$ is accumulated and weighted by each channel's density and normalized by the sum of overall densities, leading to
\begin{align} \label{equ:recombine}
\density = \sum_{i=1}^M \density^i \quad \text{and} \quad \radiance = \frac{1}{\density} \sum_{i=1}^M \density^i \radiance^i.
\end{align}
In addition to rendering all layers, it is possible to render the $i^\text{th}$ semantic layer only by using the density $\density^i$ and color $\radiance^i$ exclusively; the $i^\text{th}$ semantic \emph{mask} can be rendered by using the density $\density$ and the color $\frac{\density^i}{\density}$ as
\begin{align} \label{equ:semantic}
S^i(\ray) = \sum_{j=1}^N T_j \left(1 - \exp\left(-\density_j\delta_j\right)  \right)\frac{\density^i_j}{\density_j},
\end{align}
where $\density^i_j$ is the density for semantic layer $i$ at sample point $j$ along ray $\ray$. We only provide this abbreviated intuitive argument here for the sake of space and provide the full derivation in \suppRef{sec:compositing_derivation}. %
\Equation{nerf_composition},~\ref{equ:recombine}, and~\ref{equ:semantic} can be rigorously derived from the volume rendering equation~\cite{raytracing} by using multiple material types.

\subsection{Losses}
\label{sec:losses}
NeRF is a notoriously under-constrained representation that needs careful regularization to produce semantically correct results. 
Our layer decomposition allows each layer to be regularized independently.
During training we minimize a set of losses, each playing a critical role in the optimization process.
The full loss function is expressed as
\begin{equation}
    \loss = \loss_{\text{color}} + \lambda_{\text{sem}} \loss_{\text{sem}} + \lambda_{\text{sparse}} \loss_{\text{sparse}} + \lambda_{\text{group}} \loss_{\text{group}}, %
\end{equation}
where the $\lambda$ are hyper-parameters weighting the respective losses.
The following sections describe the loss terms.

\subsubsection{Color Loss.} We use a color loss similar to NeRF \cite{mildenhall2020nerf} to minimize the difference between the rendered and ground truth images. The loss is formulated as
\begin{align}
\loss_{\text{color}} = \sum_{\ray\in \bundle} \|C(\ray) - \hat{C}(\ray)\|^2,
\end{align}
where $\bundle$ is the set of rays in each batch, and $C(\ray)$ and $\hat{C}(\ray)$ are respectively the estimated and ground truth RGB color for the ray $\ray$.

\subsubsection{Semantic Loss.} 

In a similar spirit, we introduce a semantic loss term to minimize the difference between the rendered and ground truth semantic masks, given by
$|S^i(\ray) - \hat{S}^i(\ray)|^\gamma$, where $S^i(\ray)$ and $\hat{S}^i(\ray)$ are the $i^\text{th}$ channels of the estimated and ground truth semantic segmentation masks, respectively, along the ray $\ray$. In practice, the semantic masks used for supervision are generated from a \MaskRCNN model \cite{he2017mask} and are prone to containing outliers, hence we use a robust loss with $\gamma \leq 1$.
Additionally, to account for class imbalances, we weigh each class separately using its instantaneous recall (\ie, within each training batch)~\cite{tianStrikingRightBalance2022} leading to
\begin{align}
\loss_{\text{sem}} &= \sum_{\ray\in \bundle} \sum_{i=1}^M (1 - R^i) |S^i(\ray) - \hat{S}^i(\ray)|^\gamma, \quad \gamma = 1,
\end{align}
where $R^i  = \frac{TP^i}{P^i}$ is the recall of class $i$, i.e., the ratio of true positives $TP$ to all positives $P$ of that class.

\subsubsection{Sparsity Loss.}
Without further regularization, the reconstructed models are prone to semi-transparent material close to training camera positions and in mid-air (as will be shown in the ablation study in the experiments section).
These are over-fitting artifacts that frequently originate from incorrectly modeled viewpoint-dependent effects.
To reduce the occurrence of such partial opacities, we introduce a sparsity loss that favors solutions where opacity is trending towards 0 or 1.
This sparsity loss is defined as
\begin{align}
\loss_{\text{sparse}} = \sum_{\ray\in \bundle} \sum_{i=1}^M \sum_{j=1}^N \bigg(& |1 - \exp(-\density^i_j\delta_j)|^{\gamma} + \nonumber\\
& |\exp(-\density^i_j\delta_j)|^{\gamma} \bigg), \quad \gamma < 1.
\end{align}

\begin{figure*}[tb]%
\centering%
\includegraphics[width=\linewidth]{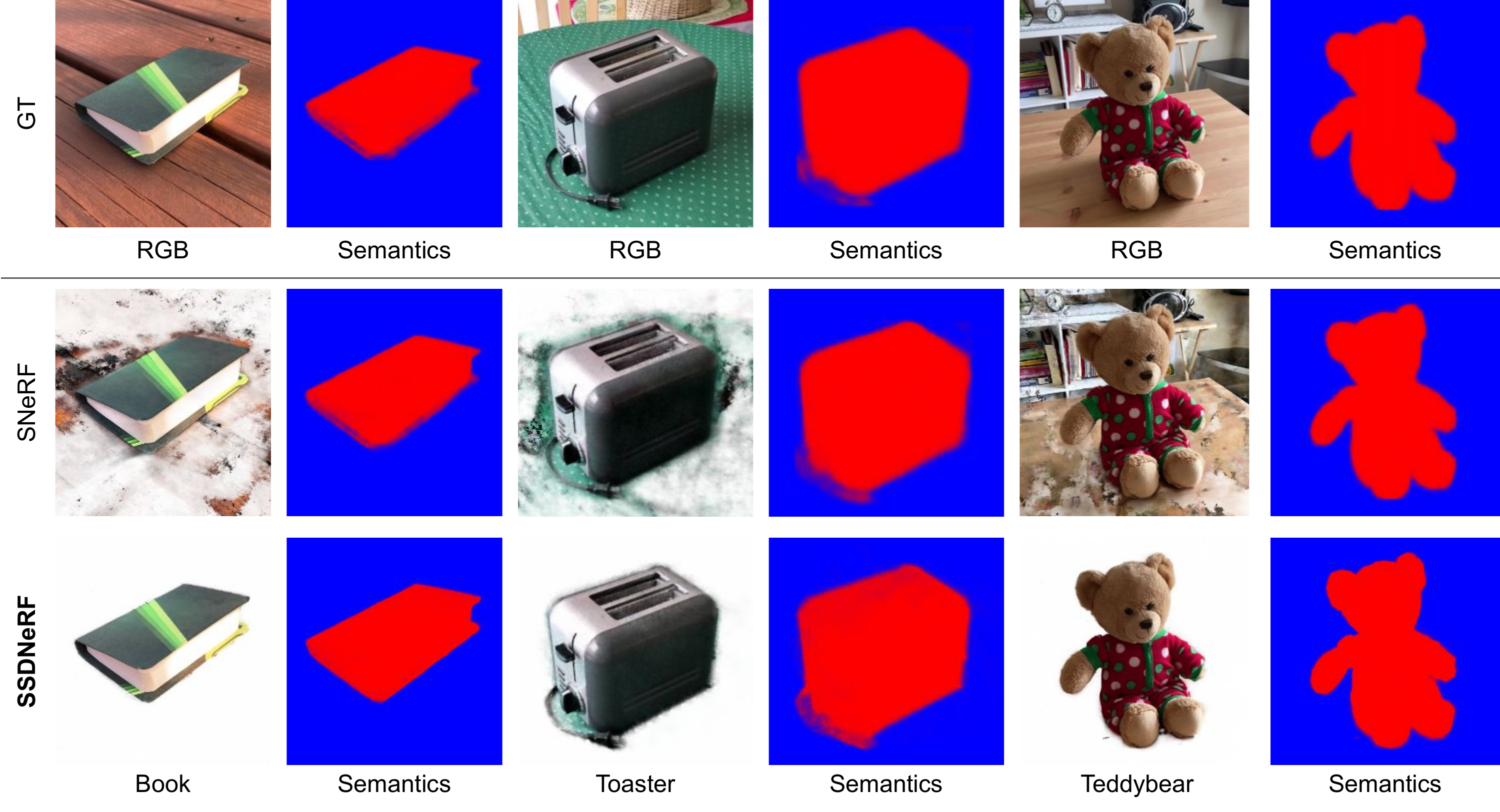}
\caption{Qualitative comparison between SNeRF and our SSDNeRF on the CO3D dataset showing reconstruction of the foreground layer and the predicted semantic probabilities. Top row shows ground-truth images and input masks. SSDNeRF achieves a much cleaner decomposition into layers.}
\label{fig:compare_snerf_co3d}
\end{figure*}

\subsubsection{Group Sparsity Loss.}
Whereas the 2D semantic segmentation masks are noisy, we do know that most of the time very few semantic classes should be present at any point in space.
We formulate this desired property in an additional regularization term minimizing the co-occurrence of opacity between semantic layers.
We use a group sparsity loss for this purpose, favoring solutions where only one opacity value is trending toward~$1$ and the others toward~$0$ for each sample, defined as
\begin{equation}
\loss_{\text{group}} = \sum_{\ray\in \bundle} \sum_{i=1}^M \sum_{j=1}^N |1-\exp(-\density^i_j\delta_j)|^{\gamma}, \quad \gamma < 1.
\end{equation}

\section{Experiments}

This section describes our experiments and presents the results of our method. We begin with the implementation details, establish comparisons with existing methods, and finally, show novel applications enabled by our method.

\def\figScale{0.15}
\setlength\tabcolsep{0.005\textwidth}
\begin{figure*}[htb]%
\centering%
\begin{tabular}{ccc|ccc}%
\includegraphics[width=\figScale\textwidth]{figs/compare_snerf/snerf/rgb_80.png}
&\includegraphics[width=\figScale\textwidth]{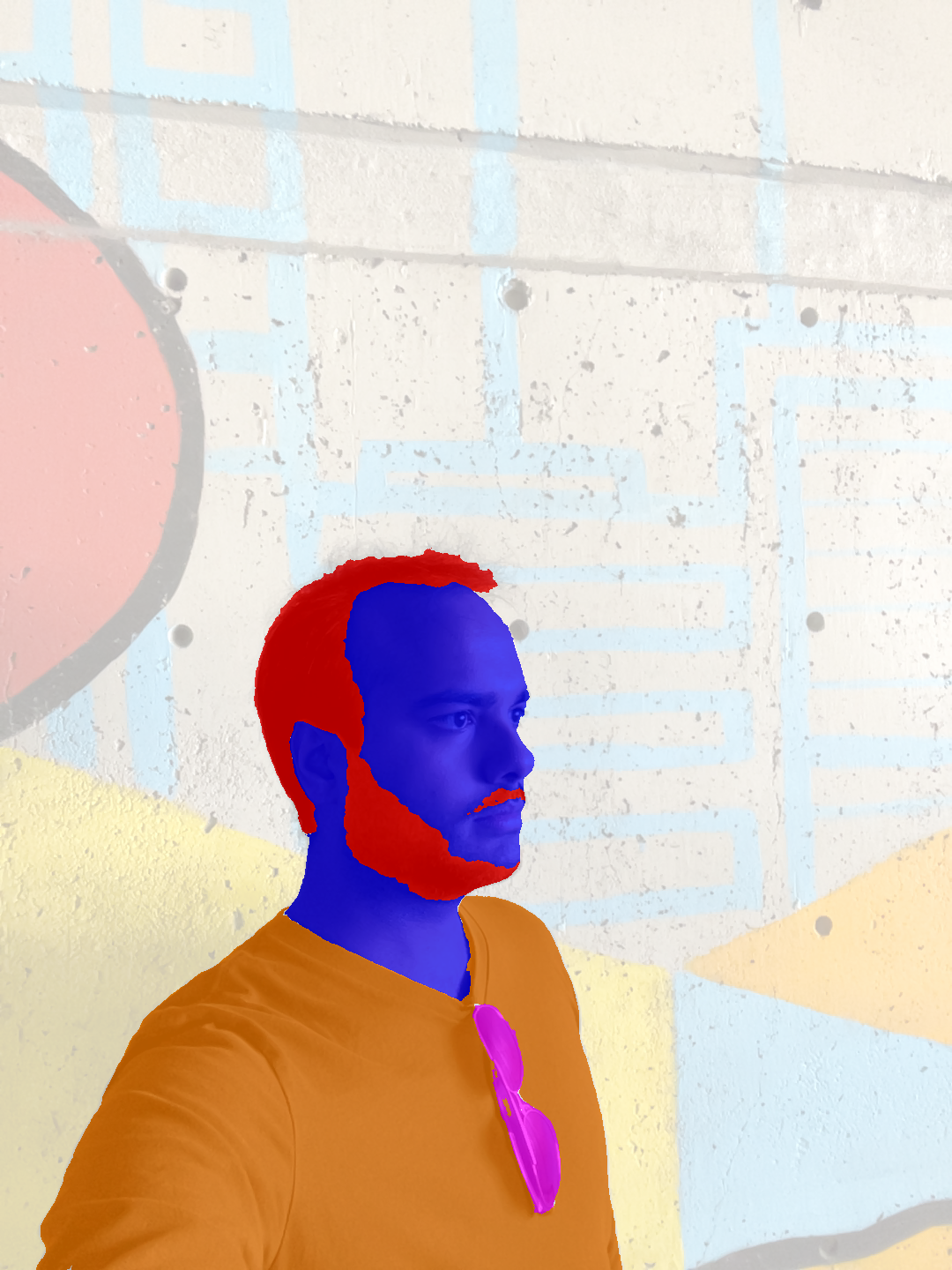}
&\adjincludegraphics[width=\figScale\textwidth,trim={{0.2\width} {0.2\height} {0.4\width} {0.4\height}},clip]{figs/compare_snerf/snerf/mask_o0.3_80.png}
& \includegraphics[width=\figScale\textwidth]{figs/compare_snerf/ours/rgb_80.png}
&\includegraphics[width=\figScale\textwidth]{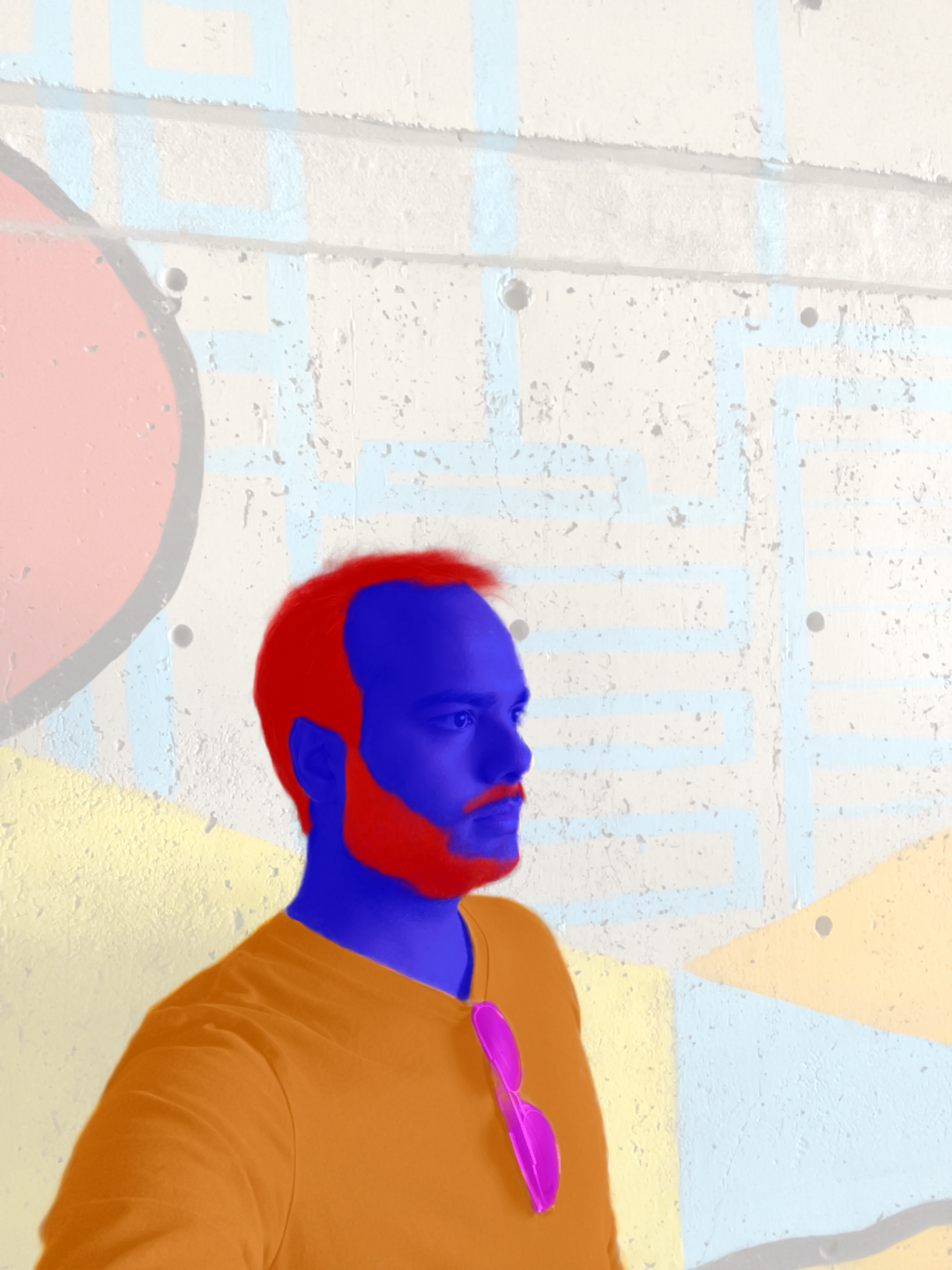}
&\adjincludegraphics[width=\figScale\textwidth,trim={{0.2\width} {0.2\height} {0.4\width} {0.4\height}},clip]{figs/compare_snerf/ours/mask_o0.3_80.png}\\
RGB & Semantics & close-up & RGB & Semantics & close-up\\
\multicolumn{3}{c|}{SNeRF~\cite{Zhi21semanticnerf}} & \multicolumn{3}{c}{\SSDNerf (ours)}
\end{tabular}
\caption{Comparison of the proposed \SSDNerf with SNeRF~\cite{Zhi21semanticnerf}, showing reconstructed RGB, semantics, and a close-up view of the semantics. SNeRF treats the semantic logits as additional radiance-like channels, and is trained with a cross-entropy loss with \emph{hard} (one-hot) labels. In contrast, \SSDNerf formulates different semantic layers as separate opacities and colors, and can correctly handle \emph{soft} labels, allowing for a decomposition of the scene into layers (\Figure{compare_snerf_layers}).}
\label{fig:compare_snerf}
\end{figure*}

\subsection{Implementation Details}
We based our implementation on the deformable NeRF framework presented in~\cite{park2021nerfies}. Our decomposition framework is fully compatible with the most recent NeRF extensions and can be trivially integrated into~\cite{torch-ngp}, a re-implementation of~\cite{mueller2022instant}. The network architecture is given in \suppRef{sec:network_arch}.
Given the input images, our approach runs in a fully automatic manner.

\subsection{Datasets}
We show results on two datasets. The first is a set of \textbf{face-capture} videos using the front camera of an iPhone 12 at a range of resolutions from $1440\times 1080$ to $3840\times 2160$. 
We filter blurry frames using the variance of the image Laplacian and use COLMAP~\cite{schoenberger2016sfm} to estimate the intrinsic and extrinsic camera parameters.
To generate the supervision for the semantic layers, we use a \MaskRCNN instance segmentation model~\cite{he2017mask} trained on a total of 39 classes. For our experiments, we use a subset of 5 classes: skin, hair, T-shirt, hoodie, and glasses. Pixels not containing any instance of the selected classes are labeled background, giving us a total of 6 classes. Not all classes are present in every capture. Due to the casual setting of the face captures, the subjects do move a little during the capture process, requiring a deformation model. 

We also show results on a variety of single-scene sequences from the \textbf{CO3D}~\cite{reizenstein21co3d} dataset. Each scene contains a single object with a background, with 100 training and validation images and semantic labels (2 classes: foreground and background).
The CO3D dataset contains static scenes, and we do not use a deformation model.

\begin{table}[b]
    \centering
    \begin{tabular}{l|c|c|c|c}
        \multirow[c]{2}{*}{\textbf{Scene}} & \multicolumn{2}{c|}{\textbf{SSDNeRF}} & \multicolumn{2}{c}{\textbf{SNeRF}} \\ \cline{2-5}
         & PSNR & mIoU & PSNR & mIoU \\ \hline
        Apple & \textbf{30.7} & \textbf{0.99} & 30.2 & 0.98 \\
        Ball & 29.0 & \textbf{0.95} & \textbf{29.9} & \textbf{0.95} \\
        Bench & 28.5 & \textbf{0.98} & \textbf{28.7} & 0.95 \\
        Book & \textbf{28.2} & \textbf{0.96} & 27.5 & 0.94 \\
        Orange & \textbf{26.0} & \textbf{0.98} & 25.7 & 0.97 \\
        Hydrant & \textbf{22.0} & \textbf{0.96} & 21.8 & 0.93 \\
        Teddybear & \textbf{29.8} & \textbf{0.99} & \textbf{29.8} & 0.97 \\
        Toaster & \textbf{18.32} & \textbf{0.91} & 18.24 & 0.90 
    \end{tabular}
    \caption{CO3D dataset: Image PSNR and segmentation mean IoU}
    \label{tab:co3d}
\end{table}

\subsubsection{Training}
We train the different models on an NVidia RTX 3080 GPU for 50k iterations of the Adam optimizer~\cite{KingmaB14} with a learning rate of $10^{-2}$, $\beta_1=0.9$, and $\beta_2=0.99$.
During training, we use 2048 random rays from an image in each batch. 
We use $\gamma = 1$ for the semantic loss, and $\gamma = 0.8$ for sparsity and group sparsity.
The weights for the losses are $\lambda_{\text{color}}=1$, $\lambda_{\text{sem}}=10^{-1}$, $\lambda_{\text{sparse}}=10^{-3}$, and $\lambda_{\text{group}}=10^{-3}$.

\subsection{Evaluation} 

To evaluate our \SSDNerf approach, we compare it with several existing state-of-the-art methods for view synthesis and (soft) segmentation. We use the exact same input, training and testing data for all the methods. We also provide a qualitative ablation study of the different losses used for training our models.

\subsubsection{\SSDNerf \vs SNeRF}%

The proposed SSDNeRF offers two key advantages over SNeRF~\cite{Zhi21semanticnerf}. %

First, \SSDNerf \emph{formulates semantics using multiple densities}, while SNeRF models semantic logits as additional radiance-like channels, composited in the same way as the color. %
This enables \SSDNerf to achieve a clean decomposition of the radiance field into layers. %
SNeRF is not built for this use-case -- we formulate a way to extract layers from it by modifying the compositing equation to remove some of the radiance contribution of 3D points based on the predicted semantic logits at those points. %
\Cref{fig:compare_snerf_layers,fig:compare_snerf_co3d} compare the decomposition of images into semantic layers of our method with SNeRF on face captures and the CO3D dataset, resp.
Because the semantic logits are unbounded, SNeRF allows low density regions to disproportionately impact the semantics. %
This results in good images and masks as long as all the layers are rendered together, since high-value logits dominate. %
However, when the contributions of some classes are removed, artifacts start becoming visible -- we refer to \suppRef{subsec:snerf_layers_supp} for further details. %

Second, \SSDNerf \emph{uses a robust loss function for supervision of the semantics}, enabling it to produce \textit{soft} labels, which are necessary to handle classes like hair. SNeRF, on the other hand, uses a cross-entropy loss, resulting in hard labels. 
\Cref{fig:compare_snerf,fig:compare_snerf_co3d} compare the semantics from our method against SNeRF~\cite{Zhi21semanticnerf} on the face captures and on the CO3D, resp.
\Cref{tab:co3d} evaluates the PSNR of the reconstruction and mean IoU of the predicted segmentation masks on several scans from the CO3D~\cite{reizenstein21co3d} dataset showing competitive performance.

The hashed-grid formulation of~\cite{mueller2022instant} used in \SSDNerf trains faster and achieves better results than NeRF~\cite{mildenhall2020nerf}, upon which the original SNeRF implementation is based. We therefore re-implement SNeRF with this formulation for fair comparison. 

\def\figScale{0.23}
\begin{figure}[htb]
     \centering
     \begin{subfigure}{\figScale\linewidth}
         \adjincludegraphics[width=\textwidth,trim={0 0 0 {0.25\height}},clip]{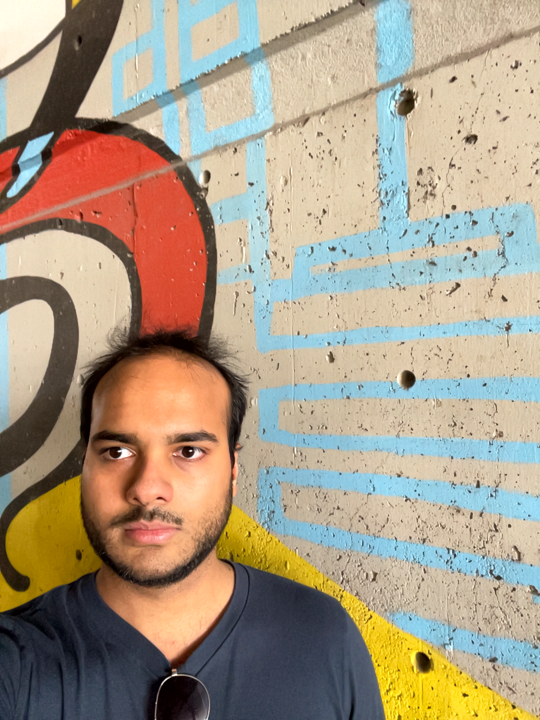}
         \caption{Input}
     \end{subfigure}
     \hfill
     \begin{subfigure}{\figScale\linewidth}
         \centering
         \adjincludegraphics[width=\textwidth,trim={0 0 0 {0.25\height}},clip,frame]{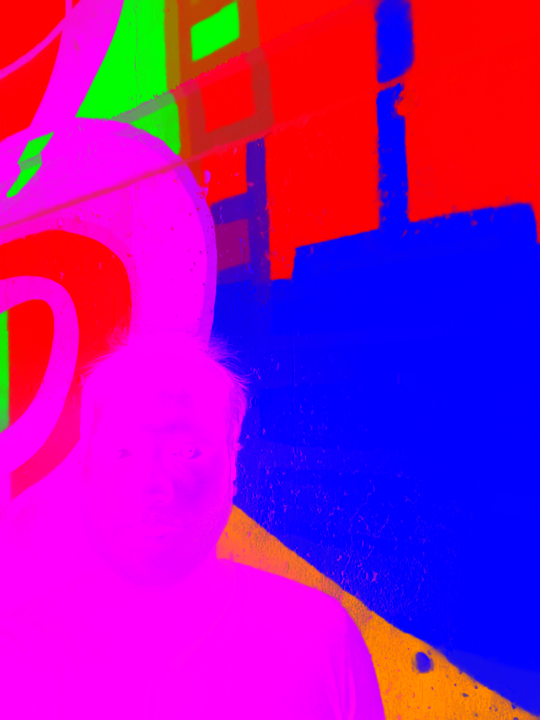}
         \caption{\cite{Levin08spectralmatting}}
     \end{subfigure}
     \hfill
     \begin{subfigure}{\figScale\linewidth}
         \centering
         \adjincludegraphics[width=\textwidth,trim={0 0 0 {0.25\height}},clip,frame]{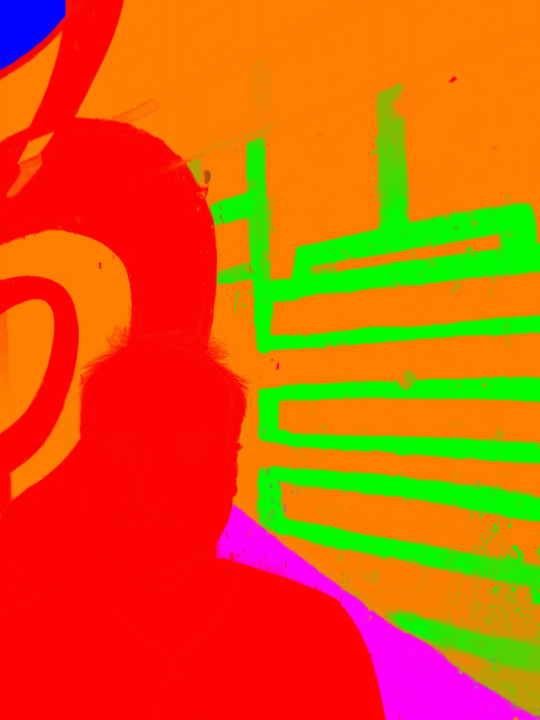}
        \caption{\cite{Aksoy2018sss}}
     \end{subfigure}
     \hfill
     \begin{subfigure}[]{\figScale\linewidth}
         \centering
         \adjincludegraphics[width=\textwidth,trim={0 0 0 {0.25\height}},clip,frame]{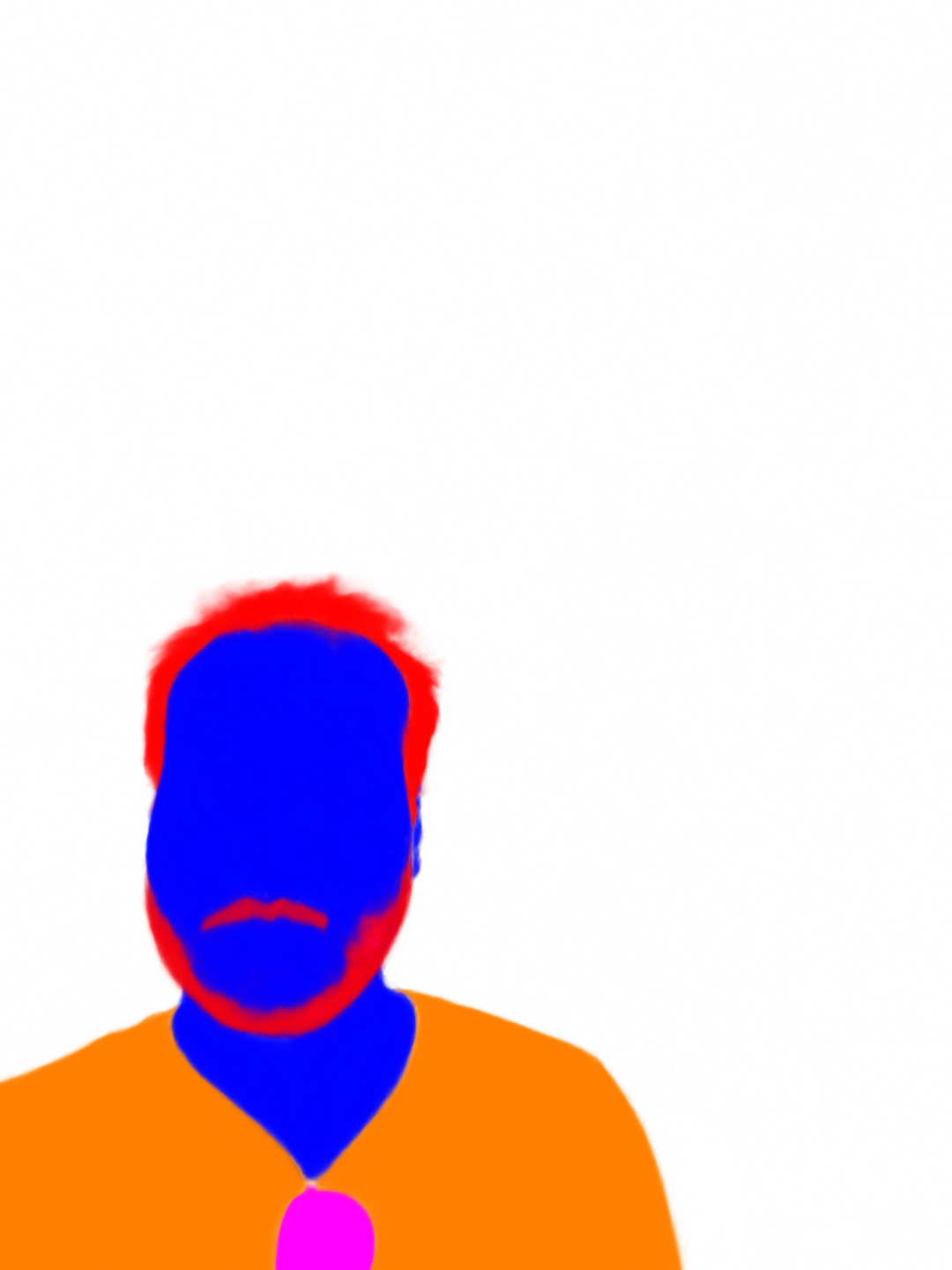}
         \caption{Ours}
     \end{subfigure}
     \vspace{-2mm}
    \caption{Comparison with existing methods: (a) input, (b) results by Spectral Matting~\cite{Levin08spectralmatting}, (c) by Semantic Soft Segmentation~\cite{Aksoy2018sss}, (d) our results. The soft segments are visualized by assigning a solid color to each segment and mixing them using alphas. Since ~\cite{Aksoy2018sss,Levin08spectralmatting} are 2D color-based, semantic classes aren't always well-preserved, resulting in inaccurate layers in contrast to our 3D approach.}
    \label{fig:comparison}
\end{figure}

\subsubsection{\SSDNerf \vs Spectral Matting Techniques}
We also compare our approach to existing soft segmentation methods. \Figure{comparison} shows a comparison with Spectral Matting~\cite{Levin08spectralmatting} and Semantic Soft Segmentation~\cite{Aksoy2018sss}. While these approaches are able to generate soft segmentation masks that respect image details, they are prone to mix different semantic classes together as they are 2D-based and use color consistency to propagate labels. Moreover, these approaches are not temporally stable and cannot be directly applied for video editing tasks (see supplementary video).  

\subsubsection{\SSDNerf vs \MaskRCNN~\cite{he2017mask}.}
\begin{figure}[tb]
    \centering
    \includegraphics[width=\linewidth]{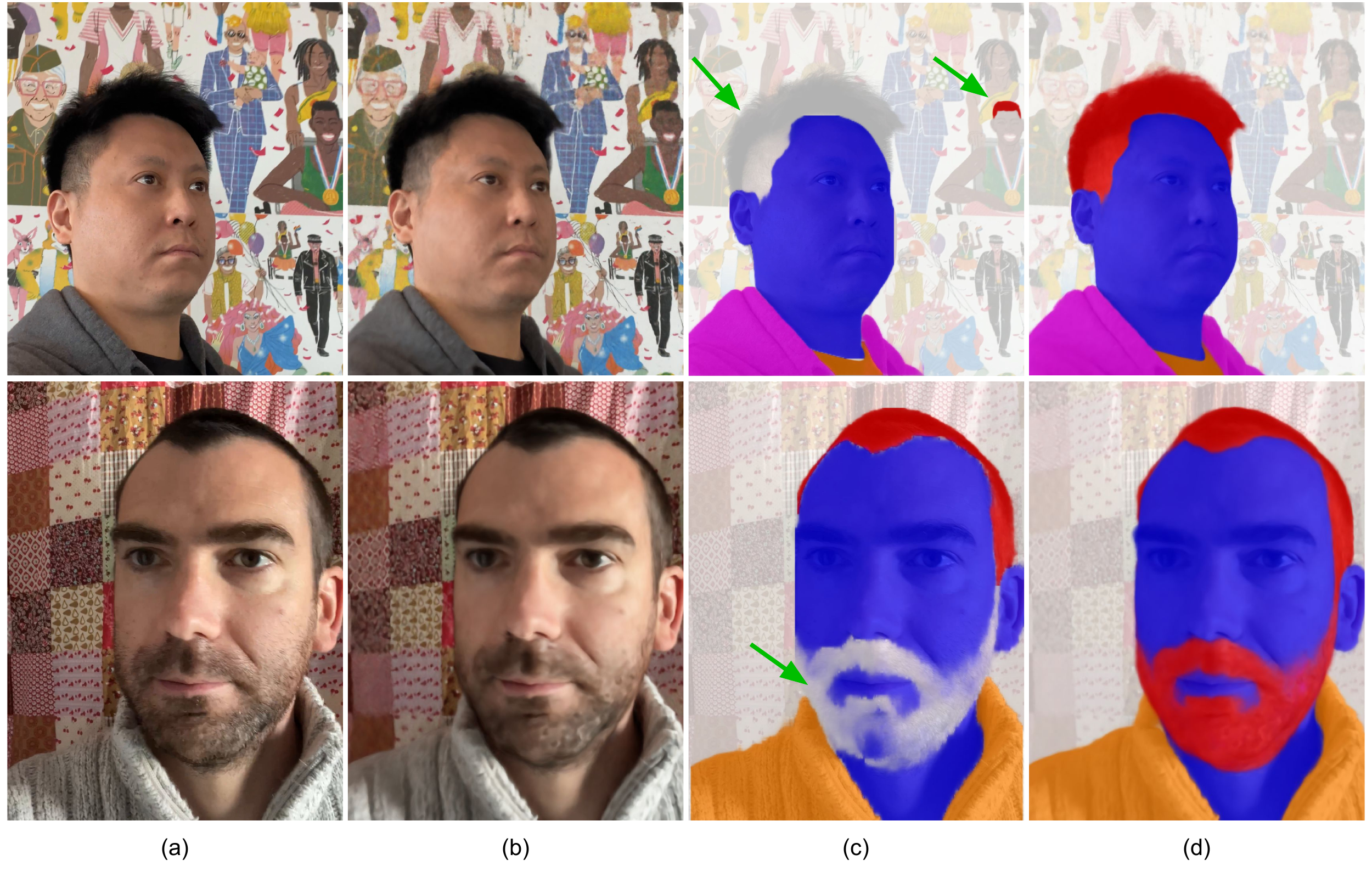} 
    \vspace{-6mm}
    \caption{Visualization of (a) input views, (b) our reconstructions, (c) semantic segmentations from \MaskRCNN (used as input), (d) predicted semantics by our approach. Our approach aggregates the semantic segmentation results from \emph{multiple} views. This allows to handle segmentation outliers: see the incorrect segmentations shown by the green arrows in column (c) and how they are robustly recovered in our results in column (d). This allows to produce soft segmentation masks that are temporally stable (see supp.\ video).}
    \label{fig:mask_vis}
\end{figure}

In \Figure{mask_vis}, we compare our results with the semantic masks generated by \MaskRCNN. By consolidating the \MaskRCNN labels using multiple views, our approach generates outlier-free results as well as soft boundaries.

\subsubsection{Qualitative ablation study}
\begin{figure}[t!]
    \centering
    \includegraphics[width=\linewidth]{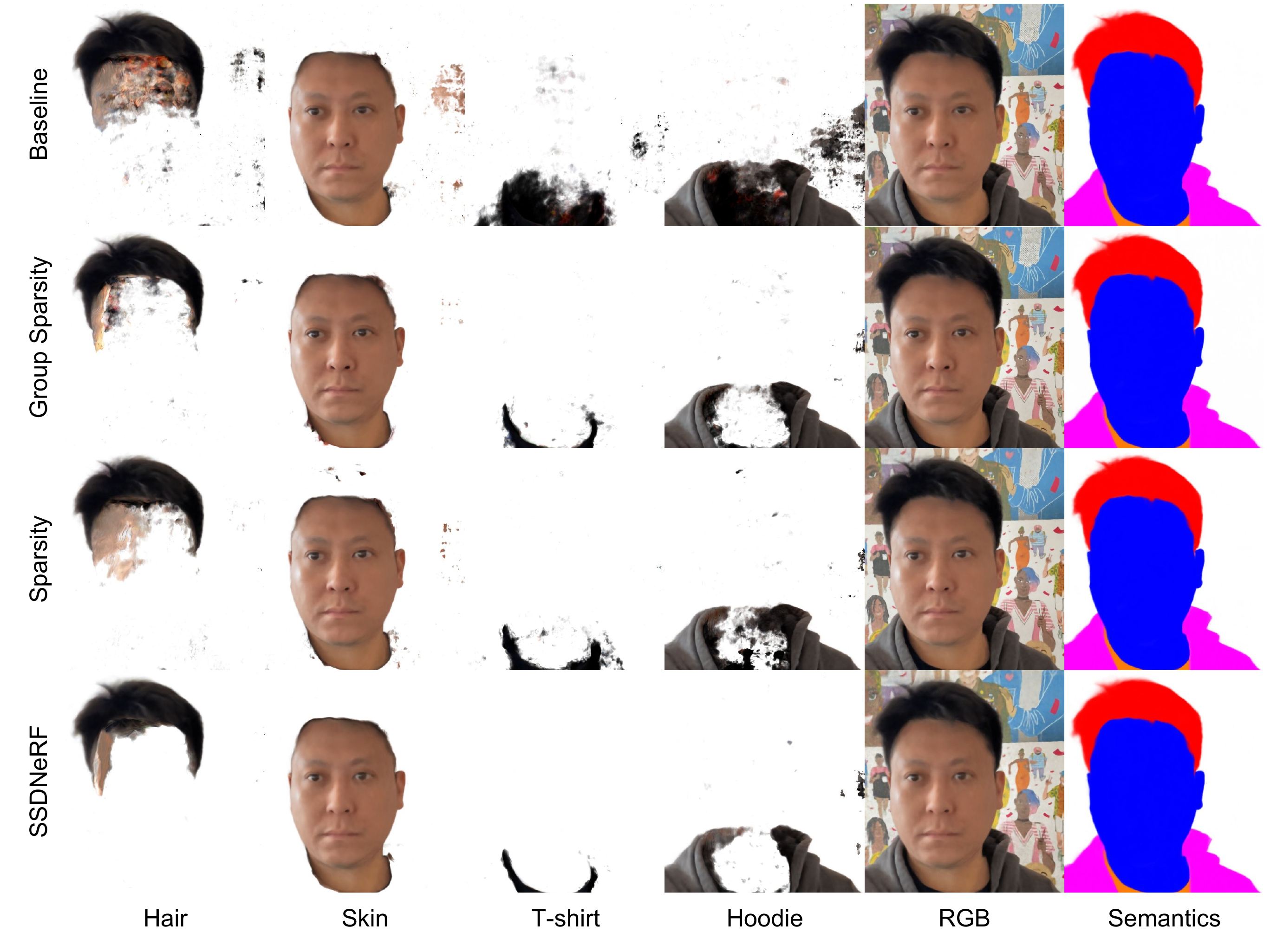}
    \vspace{-7mm}
    \caption{Ablation of the proposed regularizers: adding the regularization terms improves the reconstruction quality of each of the individual layers (see hair, skin, T-shirt, and hoodie columns), without adversely affecting the reconstructed image and semantics (see RGB and semantic columns).}
    \label{fig:ablation}
\end{figure}

We show the effect of the proposed regularizers in \Figure{ablation}. The baseline (with only the color and semantic losses) produces an appropriate RGB and semantic reconstruction, but the individual semantic layers contain \inquote{floater} artifacts. These floaters are occluded by other visible parts of the scene when rendering the entire scene, but become visible when rendering individual layers.
The proposed regularizers significantly reduce these artifacts, and the final \SSDNerf result (with both regularizers, last row) produces better results than the baseline and those obtained using each regularizer separately.

\begin{figure}[hb!]
\centering%
\def\hspaceVal{0.002} 
\def\hspaceValAfter{0.002} 
\def\figScale{0.20}
\begin{tabular}{c|ccc}
Original colors & \multicolumn{3}{c}{Temporally consistent edit} \\ %
\adjincludegraphics[width=\figScale\linewidth,trim={0 0 0 {0.15\height}},clip]{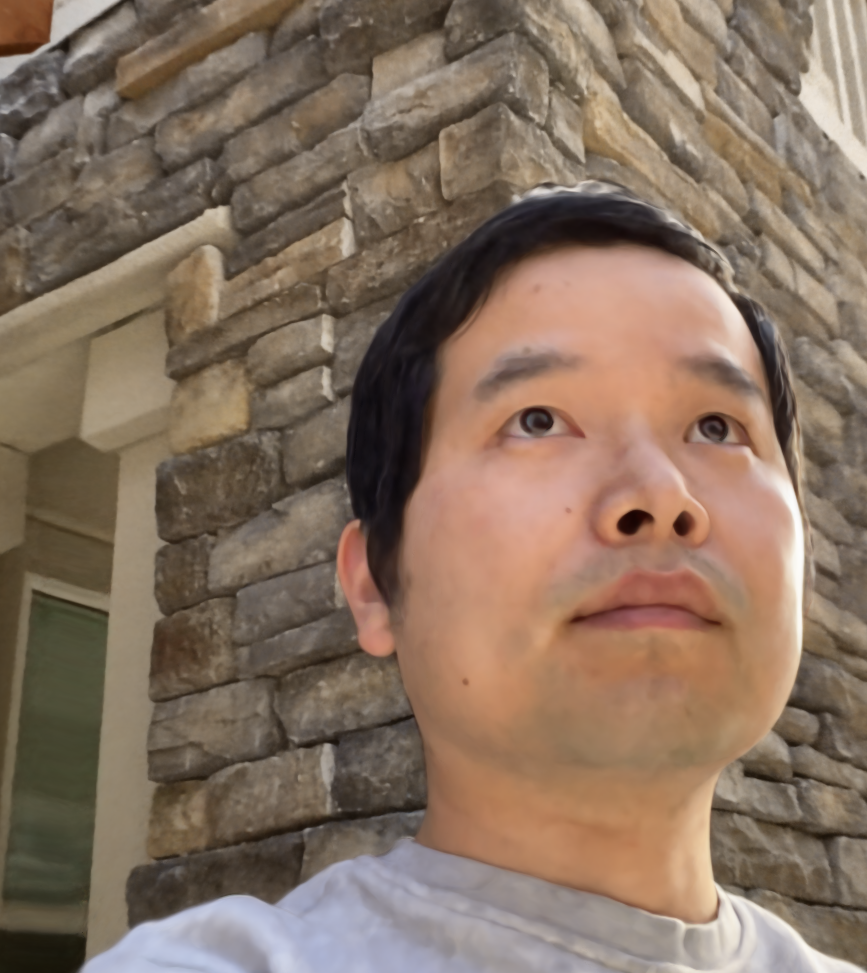}%
\hspace{\hspaceVal\linewidth}
& \hspace{\hspaceValAfter\linewidth} \adjincludegraphics[width=\figScale\linewidth,trim={0 0 0 {0.15\height}},clip]{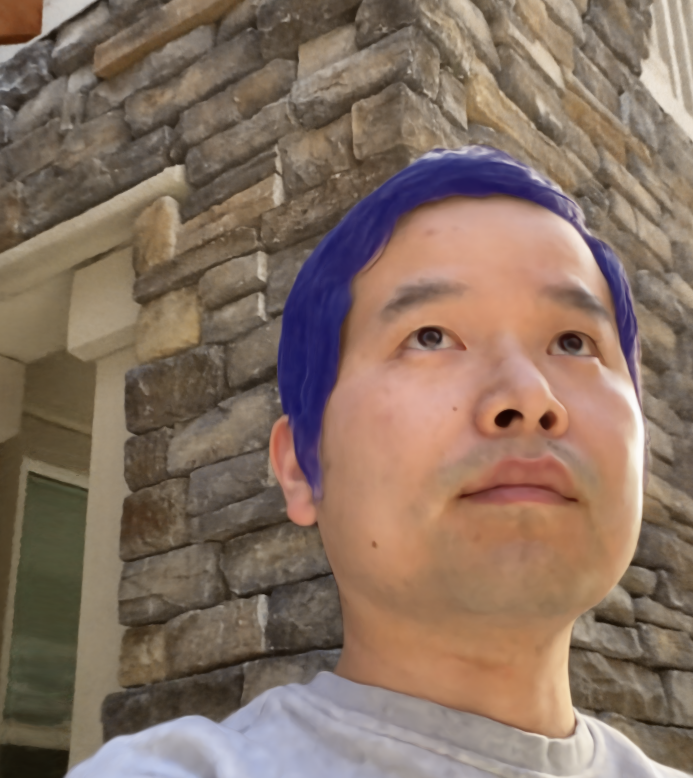}%
&\adjincludegraphics[width=\figScale\linewidth,trim={0 0 0 {0.15\height}},clip]{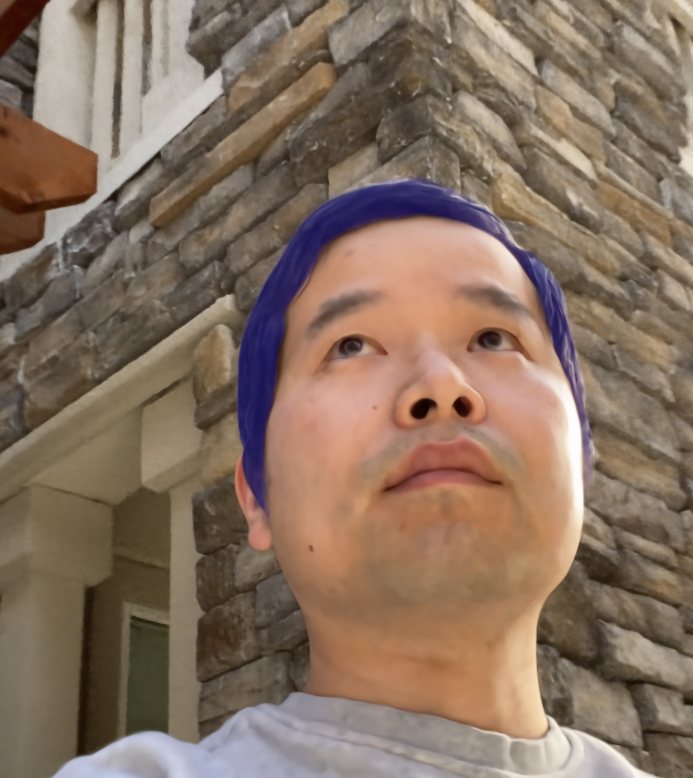}%
&\adjincludegraphics[width=\figScale\linewidth,trim={0 0 0 {0.15\height}},clip]{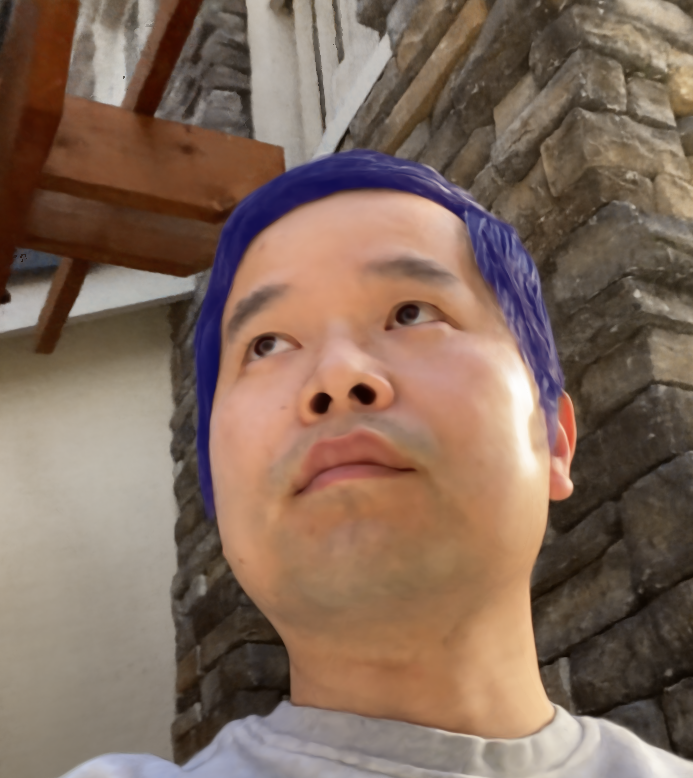} \\
\includegraphics[width=\figScale\linewidth]{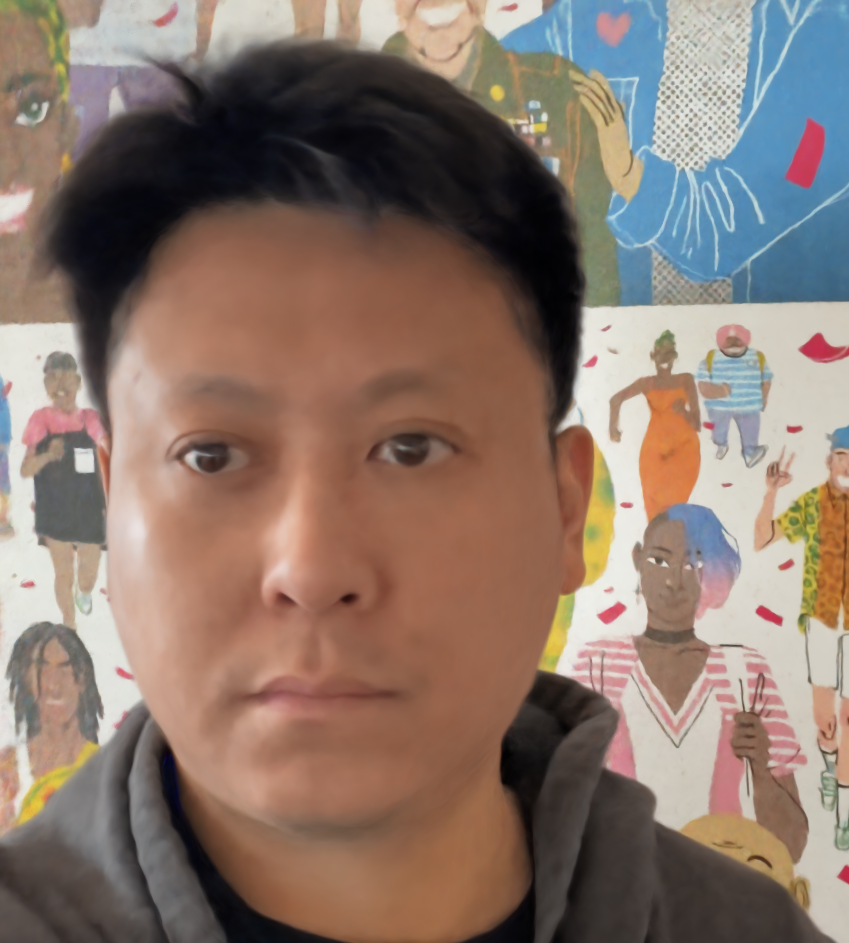}%
\hspace{\hspaceVal\linewidth}
& \hspace{0.007\linewidth}\includegraphics[width=\figScale\linewidth]{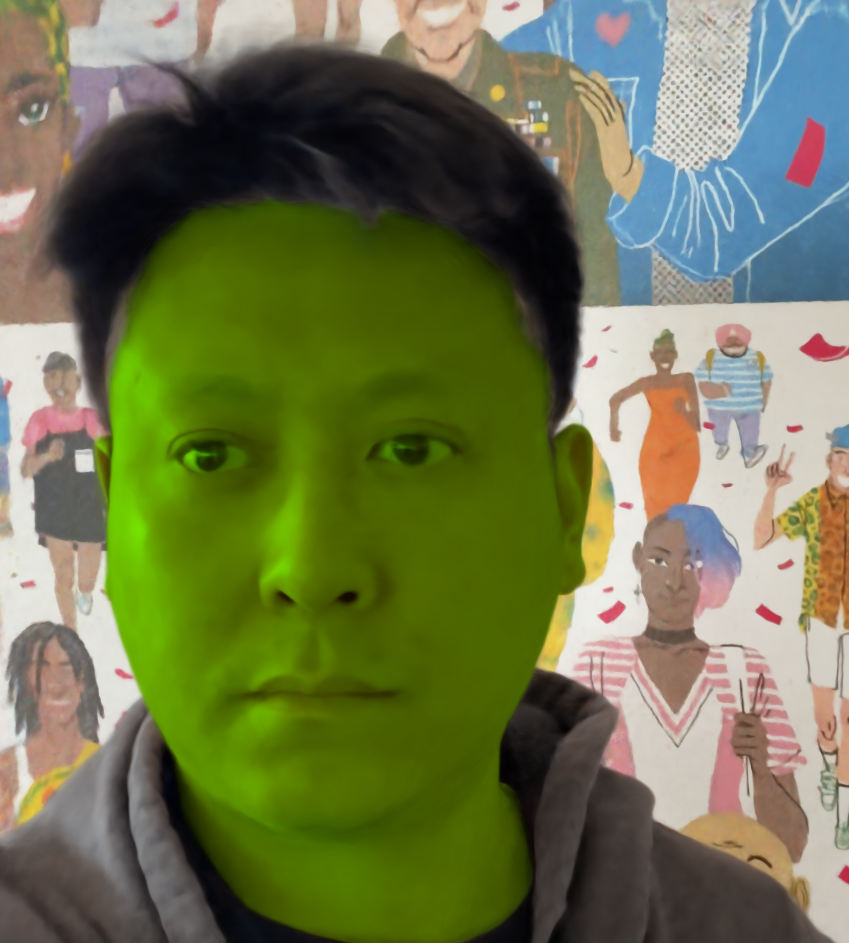}%
&\includegraphics[width=\figScale\linewidth]{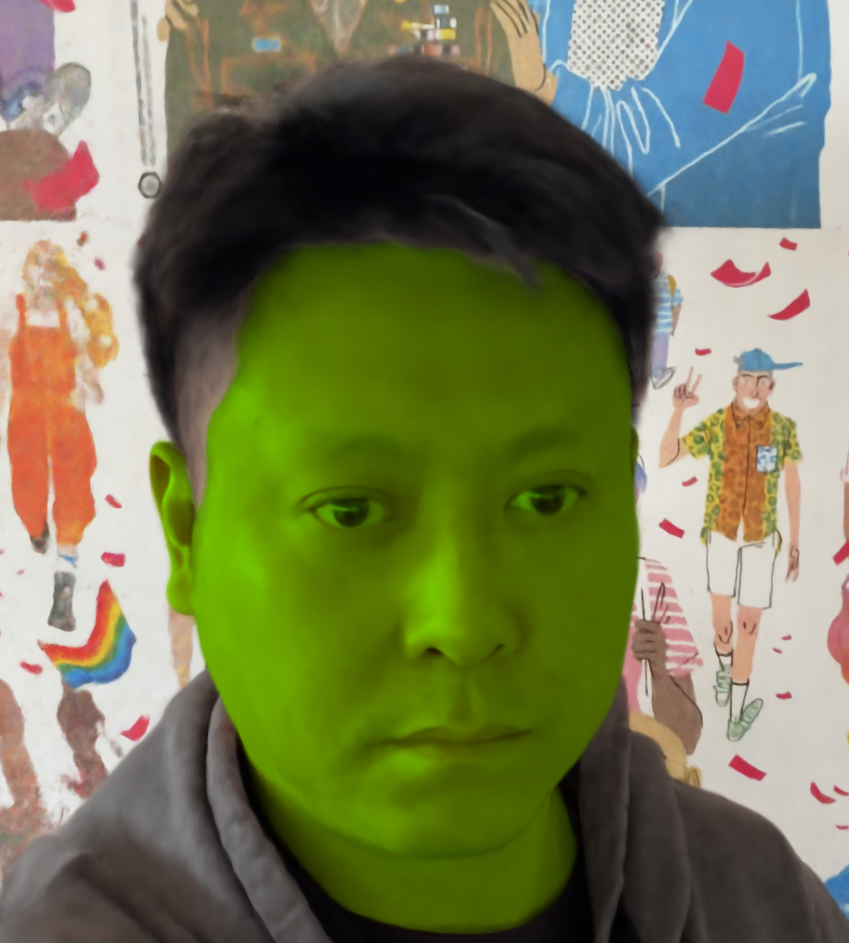}%
&\includegraphics[width=\figScale\linewidth]{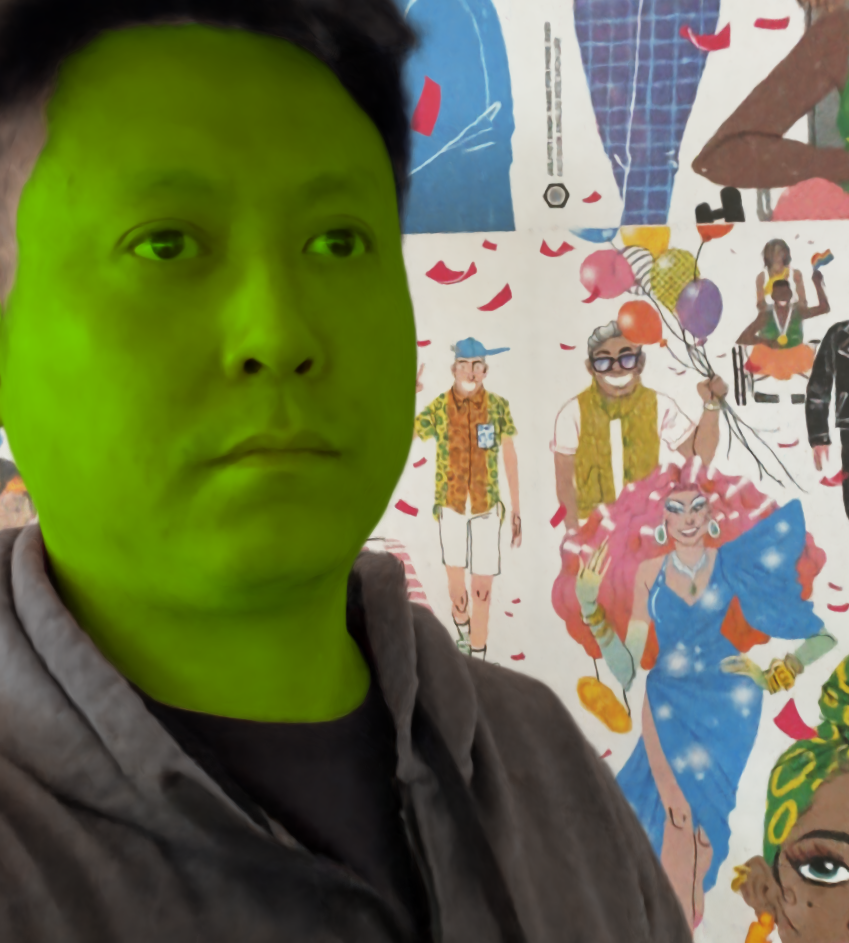} \\
\adjincludegraphics[width=\figScale\linewidth,trim={0 {0.15\height} 0 {0.15\height}},clip]{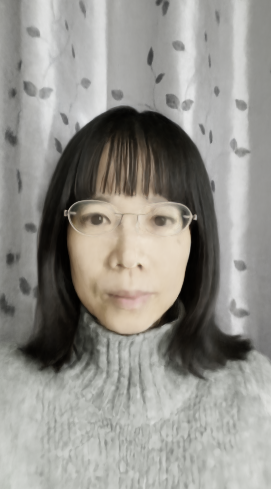}%
\hspace{\hspaceVal\linewidth}
& \hspace{\hspaceVal\linewidth}\adjincludegraphics[width=\figScale\linewidth,trim={0 {0.15\height} 0 {0.15\height}},clip]{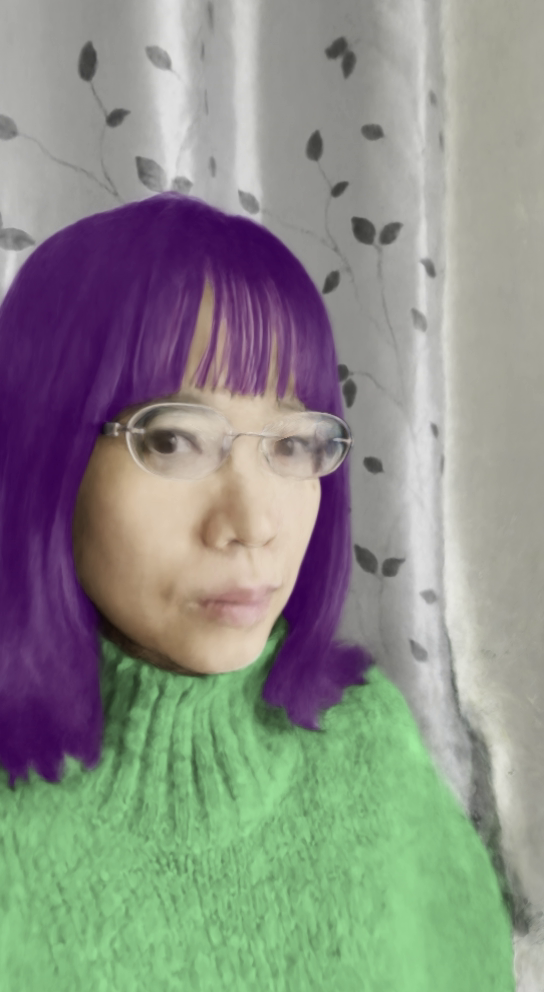}%
&\adjincludegraphics[width=\figScale\linewidth,trim={0 {0.15\height} 0 {0.15\height}},clip]{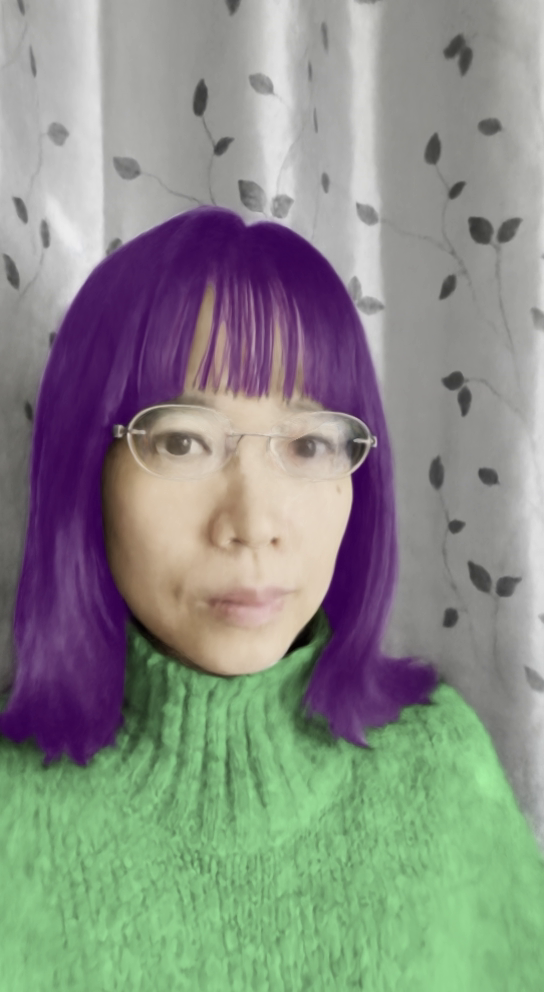}%
&\adjincludegraphics[width=\figScale\linewidth,trim={0 {0.15\height} 0 {0.15\height}},clip]{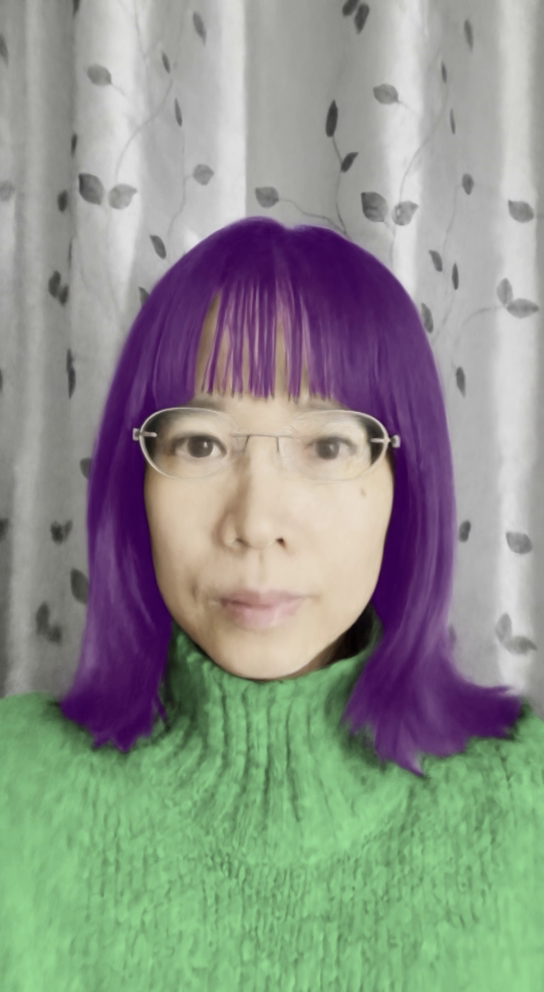} \\
\adjincludegraphics[width=\figScale\linewidth,trim={0 {0.3\height} 0 0},clip]{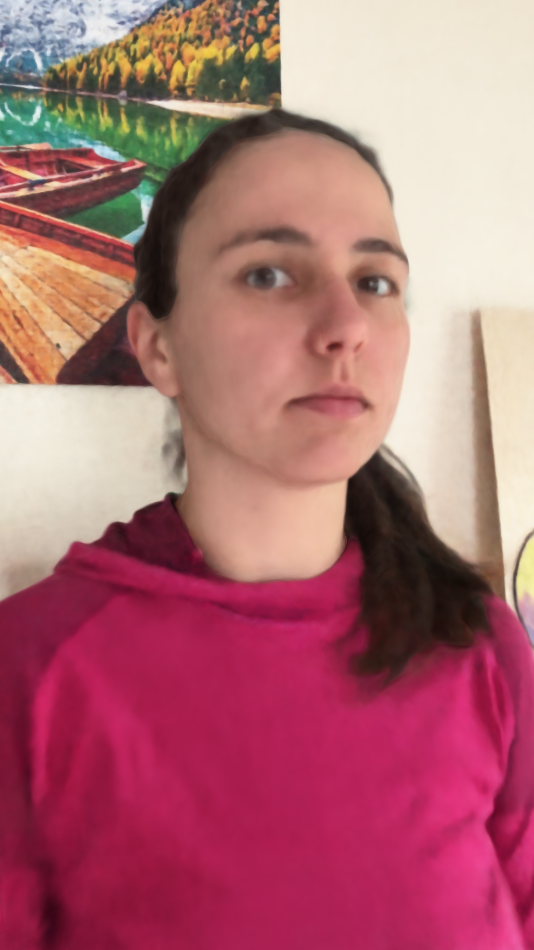}%
\hspace{\hspaceVal\linewidth}
& \hspace{\hspaceVal\linewidth}\adjincludegraphics[width=\figScale\linewidth,trim={0 {0.3\height} 0 0},clip]{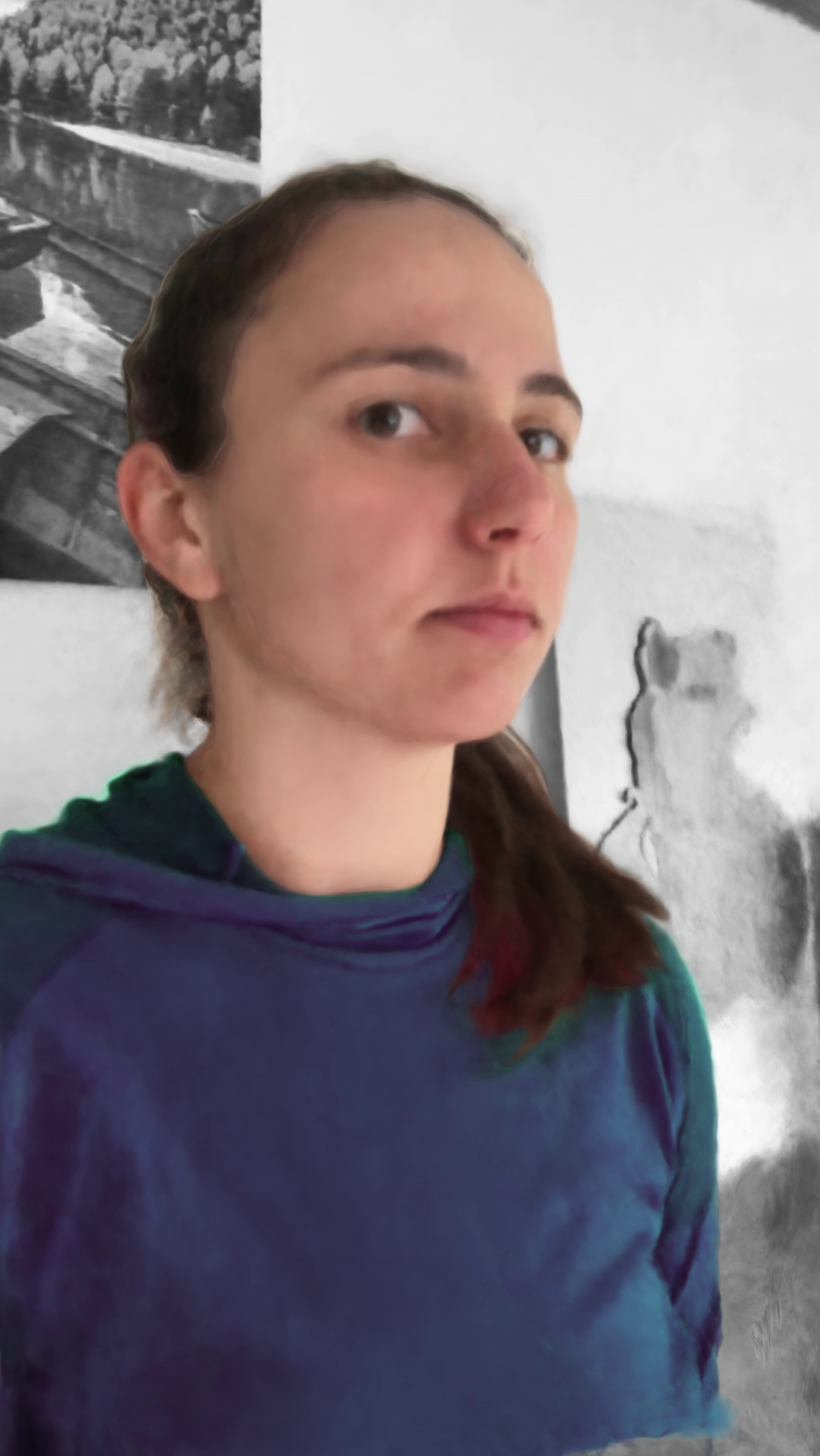}%
&\adjincludegraphics[width=\figScale\linewidth,trim={0 {0.3\height} 0 0},clip]{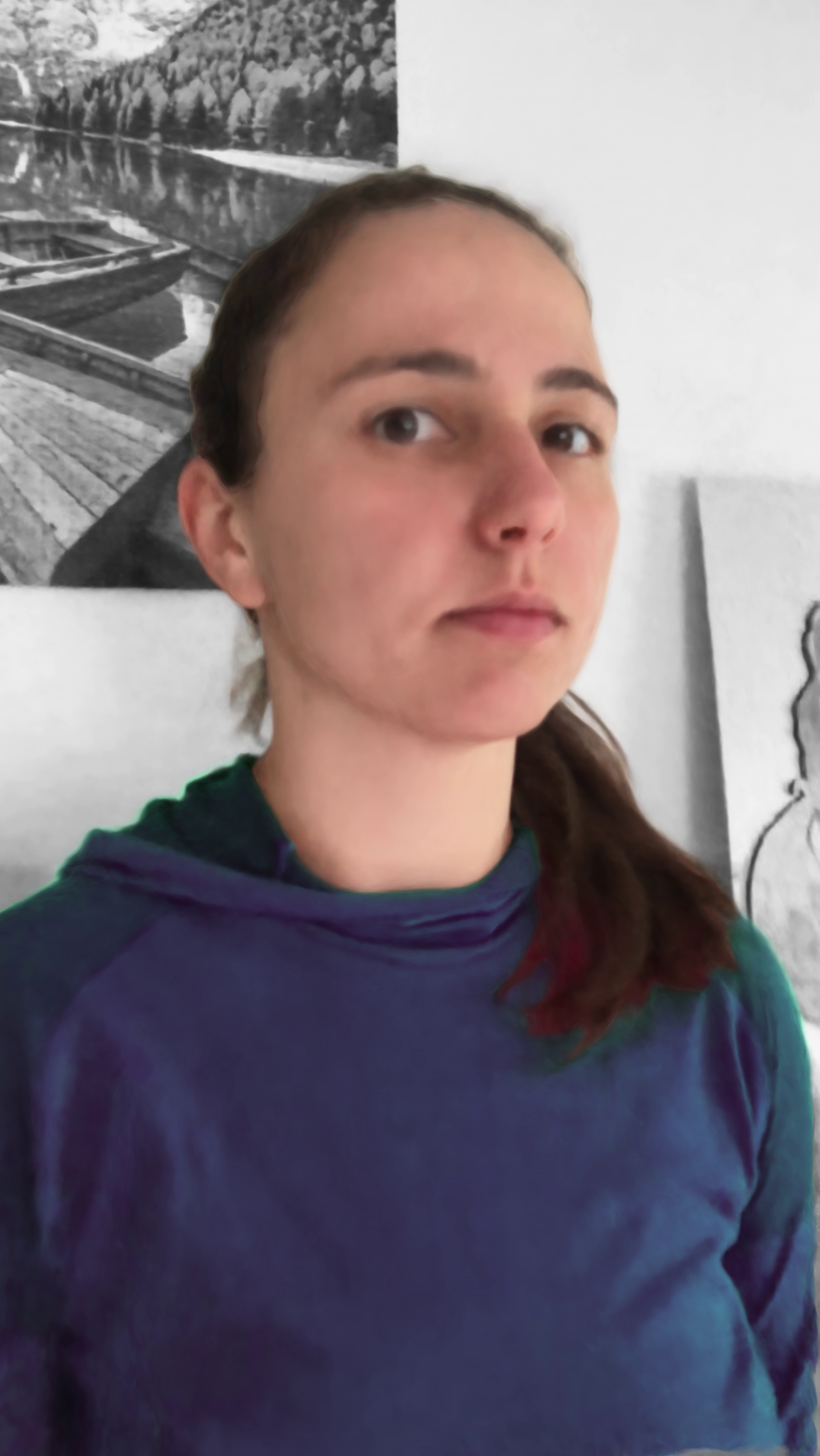}%
&\adjincludegraphics[width=\figScale\linewidth,trim={0 {0.3\height} 0 0},clip]{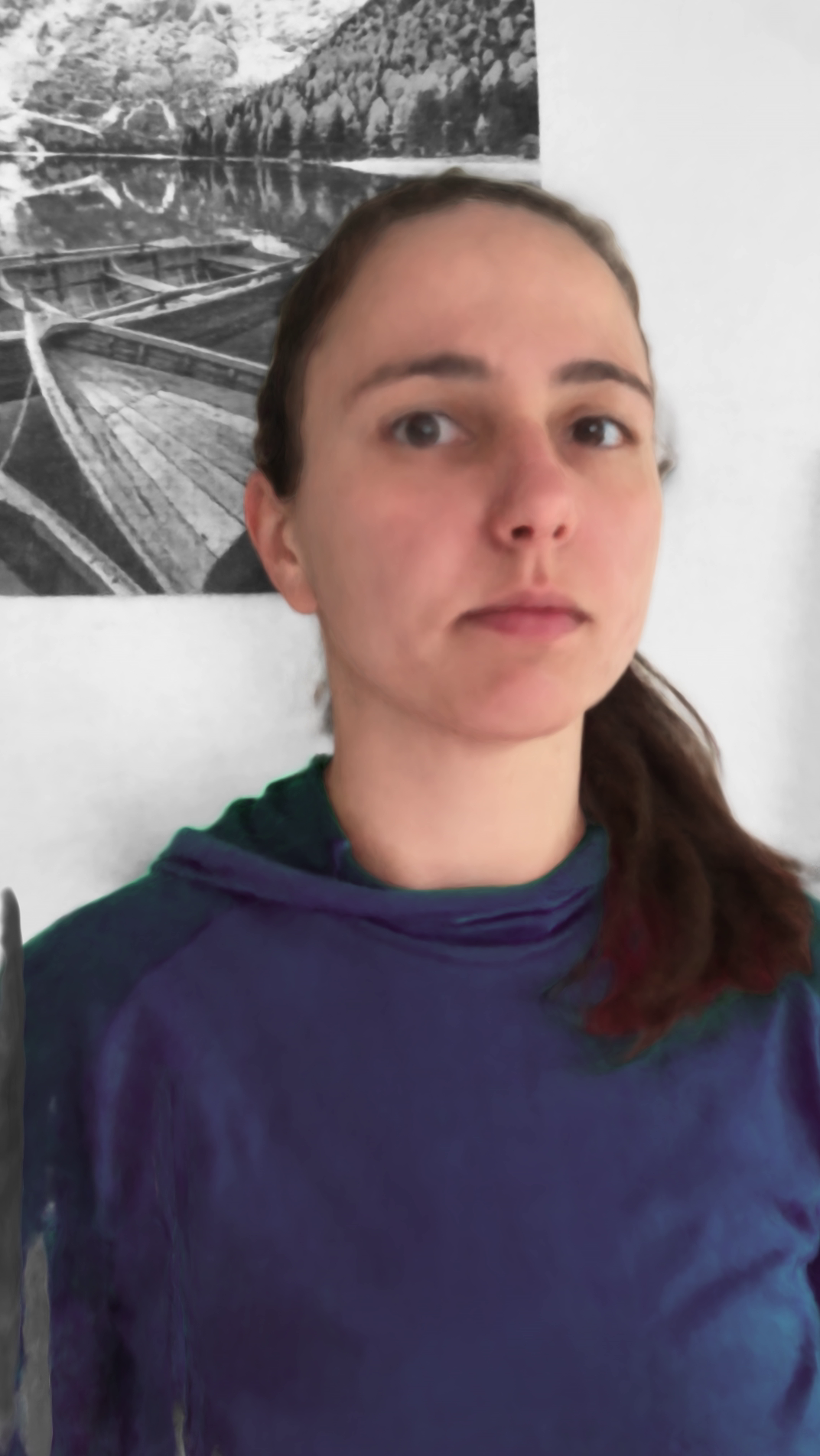}%
\end{tabular}
\vspace{-4mm}
    \caption{Examples of temporally consistent video editing. Using the semantics allows us to trivially re-color all pixels of a particular class in a temporally and spatially consistent manner without affecting others.}
    \label{fig:editing}
\end{figure}
\subsection{Video Editing Results}
In this section, we show that our approach enables several video editing applications and produces photorealistic results that are temporally and spatially stable. We demonstrate three types of edits: \Circle{1} appearance manipulation; \Circle{2} geometry manipulation; and \Circle{3} camera manipulation.

\subsubsection{Appearance Editing.}
Our extracted soft semantic layers allow us to trivially re-color all pixels of a particular class without affecting others.
\Figure{editing} shows examples of temporally consistent video edits where we transform the color of different classes. %

\begin{figure}[t!]
    \includegraphics[width=0.99\linewidth]{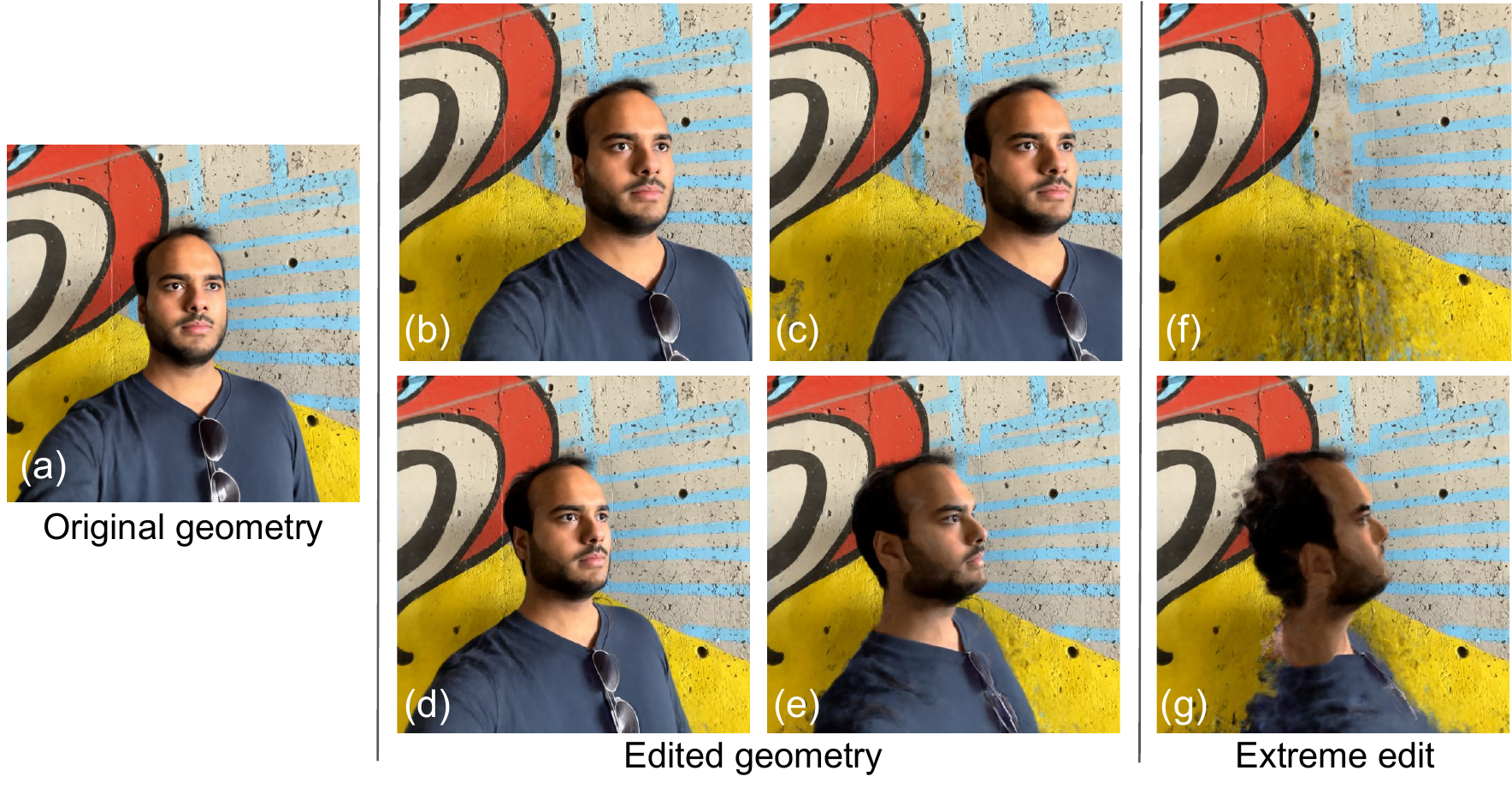}%
    \vspace{-2mm}
    \caption{Geometry editing: the semantic layers produced by our method can be independently modified geometrically without affecting other layers. We translate (top row) and rotate (bottom row) the geometry belonging to the hair, skin, shirt, and glasses layers while keeping the background fixed. 
    The rightmost column shows extreme edits. Translating by a large distance can remove the object from the view frustum entirely (f). Extreme edits have limitations, for example, rotating by too large angles brings the back of the head into the view, but since it was not visible in the training views, the head looks incomplete (g).
    }
    \label{fig:geom_edit_rot}
\end{figure}

\subsubsection{Geometry Editing.}
One benefit of the compositional nature of our approach is the ability to manipulate objects \emph{in 3D}. In \Figure{geom_edit_rot} we demonstrate that objects can be geometrically manipulated (here with a rigid 3D transformation), or even completely removed from the scene. We achieve this by sampling the radiance field a second time at the transformed coordinates, using the original samples of the background class, and the transformed samples for the foreground classes (hair, skin, shirt, and glasses), before compositing along the ray. Visually speaking, this provides novel images where the 3D pose of the person is modified, while the background is kept unchanged. Artifacts show up when the parts of the scene that are not seen in the training views (such as the back of the head in the rightmost column) become visible.

\subsubsection{Camera Editing.}
Editing extrinsic and intrinsic camera parameters can be easily achieved with NeRF approaches. This allows to generate interesting cinematographic effects such as the dolly zoom effect (see \Figure{camera_fx}). This effect is best demonstrated using video; in the supplementary video, we show that our approach can generate photorealistic novel views and reconstruct the training views with high PSNR.

\begin{figure}[t!]
    \centering
    \def\figW{0.245\linewidth}
    \adjincludegraphics[width=\figW,trim={0 0 0 {0.25\height}},clip]{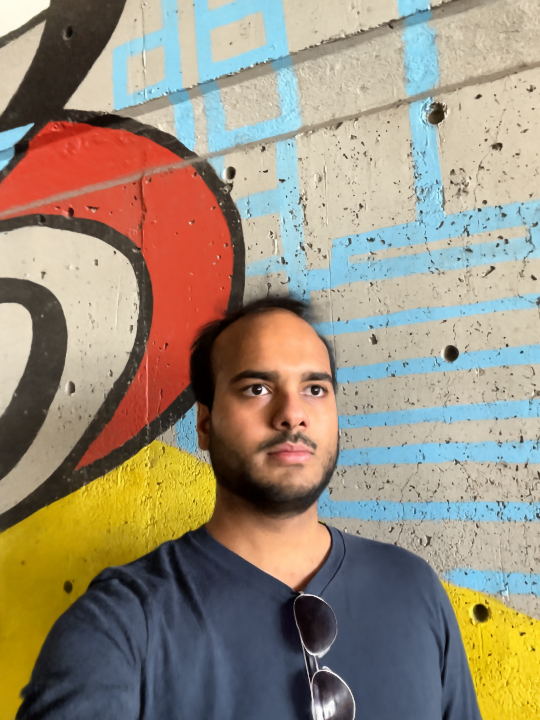}\hfill%
    \adjincludegraphics[width=\figW,trim={0 0 0 {0.25\height}},clip]{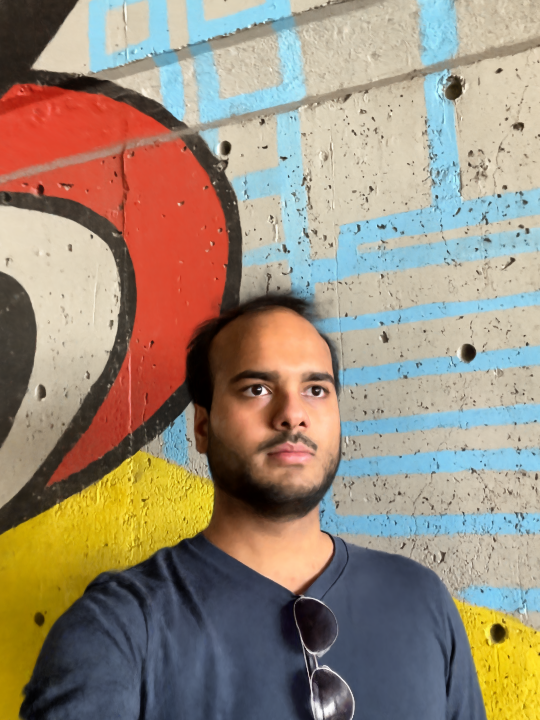}\hfill%
    \adjincludegraphics[width=\figW,trim={0 0 0 {0.25\height}},clip]{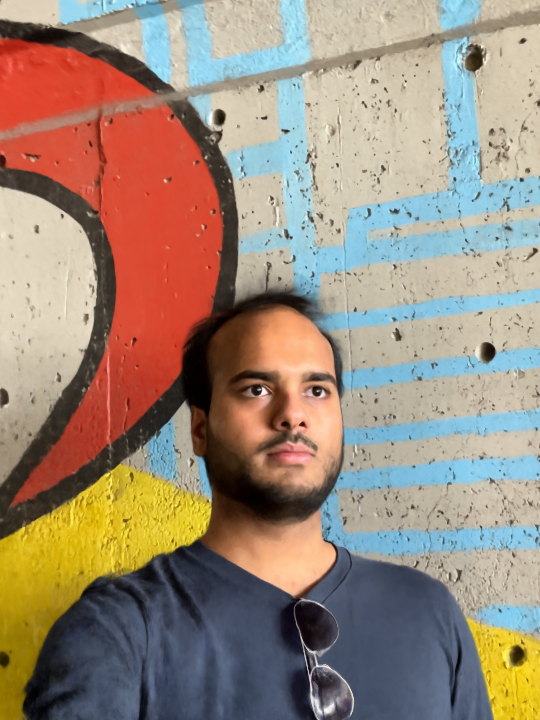}\hfill%
    \adjincludegraphics[width=\figW,trim={0 0 0 {0.25\height}},clip]{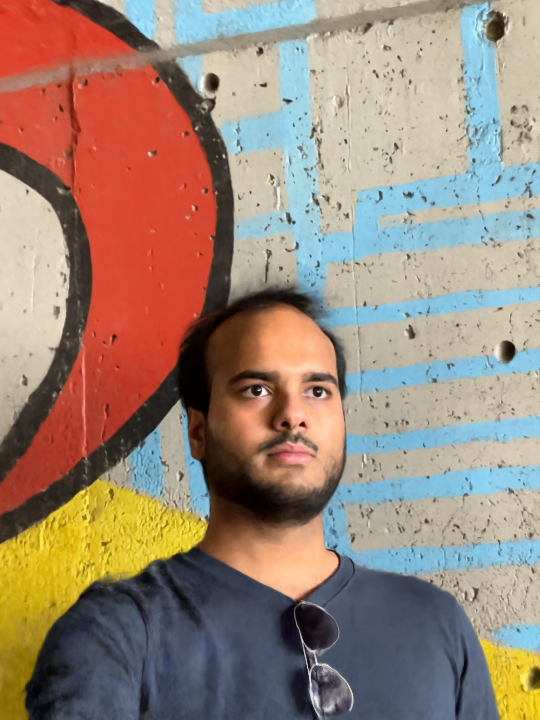}\hfill
    \caption{Dolly zoom effect: the camera simultaneously moves towards the object and reduces its focal length to keep the object size fixed, causing the background to appear to move}
    \label{fig:camera_fx}
\end{figure}
\section{Limitations and Future Work}

Our approach relies on semantic masks generated by \MaskRCNN~\cite{he2017mask}. 
While our approach is fairly robust to outliers, as demonstrated in Figure~\ref{fig:mask_vis}, the results might degrade for challenging scenes where \MaskRCNN consistently predicts wrong classes. 
One avenue of future work would be to reduce the dependency of our approach on \MaskRCNN and train it in an unsupervised manner, similar to \cite{Sbai20unsuperdecomp}. 
While we have focused our work on encoding semantic segmentation masks, other interesting semantic information could be consolidated in the same way, such as keypoint heatmaps~\cite{isack2020repose}, UV maps~\cite{guler2018densepose}, or NOX maps~\cite{sridhar2019multiview}, creating videos with a rich set of temporally and spatially consistent labels.
\section{Conclusion}
We present a new approach for representing 3D scenes with multiple encoded classes, estimating density and radiance fields for each semantic layer. We leverage a supervised training scheme with initial 2D segmentation masks from a \MaskRCNN model to train such a Semantic Soft Decomposition of Neural Radiance Fields (\SSDNerf). More importantly, we introduce the sparsity and group sparsity losses to avoid a semi-transparent material and reduce the cross-talk between layers, respectively. Experimental results demonstrate the effectiveness and superiority of the proposed \SSDNerf, compared to previous segmentation and NeRF methods. Furthermore, our approach enables 3D scene editing or re-composition, such as changing the background, object re-colorization, and geometric manipulation, in a temporally consistent manner and from novel viewpoints. Finally, we believe that the ability of our approach to generate photorealistic camera, appearance and geometric manipulations will open the door to the capture and generation of a large dataset of dynamic scenes, containing RGB images, depth maps as well as a rich set of semantic labels.

\appendix
\section{Network Architecture}\label{sec:network_arch}

\begin{figure}[h!]
    \centering
    \includegraphics[width=\linewidth]{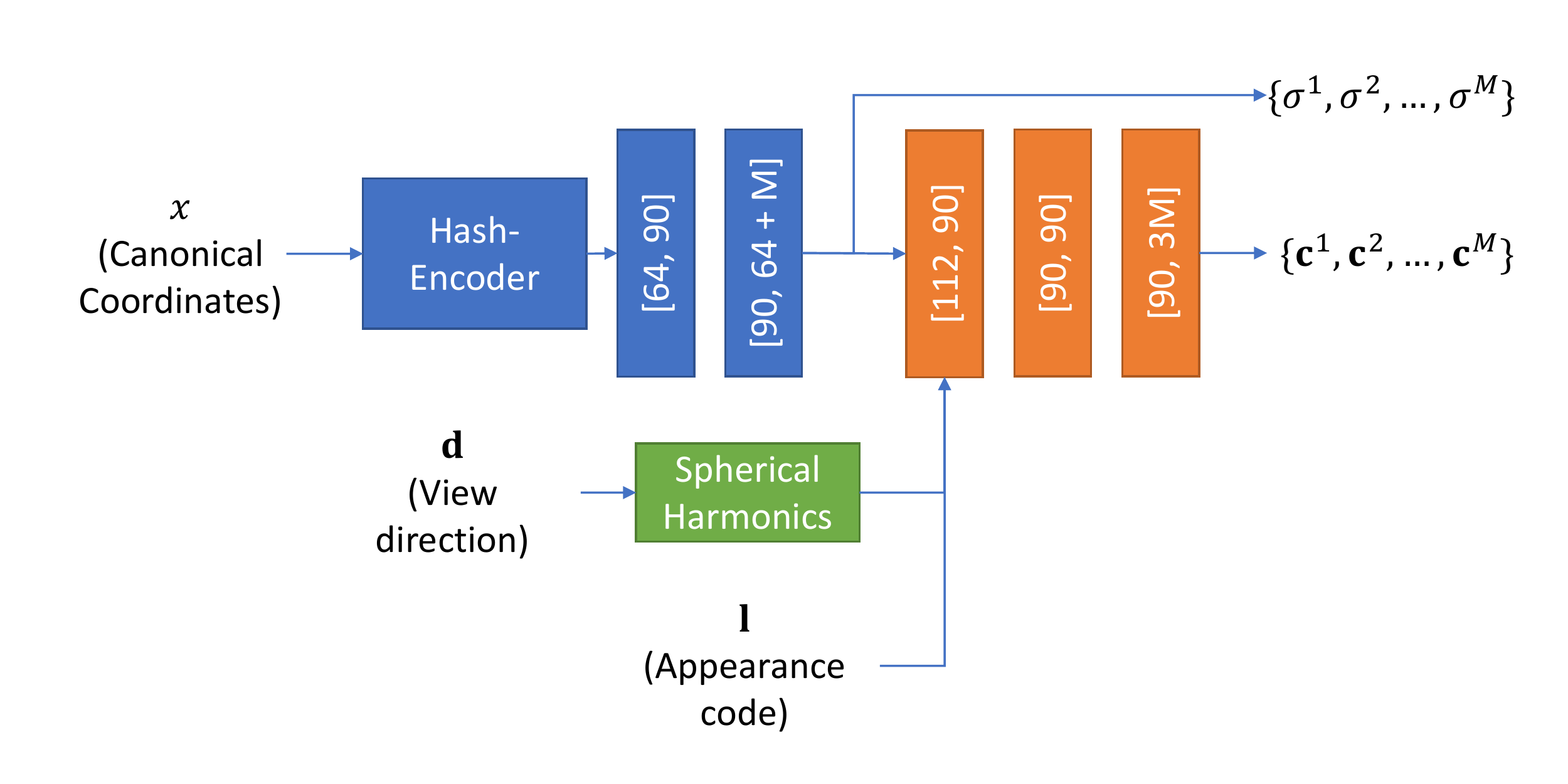}
    \caption{\SSDNerf architecture: at each point $\point$ in the canonical frame with view direction $\direction$ and appearance code $\latent$, our method predicts colors $\radiance^i$ and opacities $\density^i$ for each semantic class.}
    \label{fig:arch_new}
\end{figure}

\begin{figure}[h!]
    \centering
    \includegraphics[width=\linewidth]{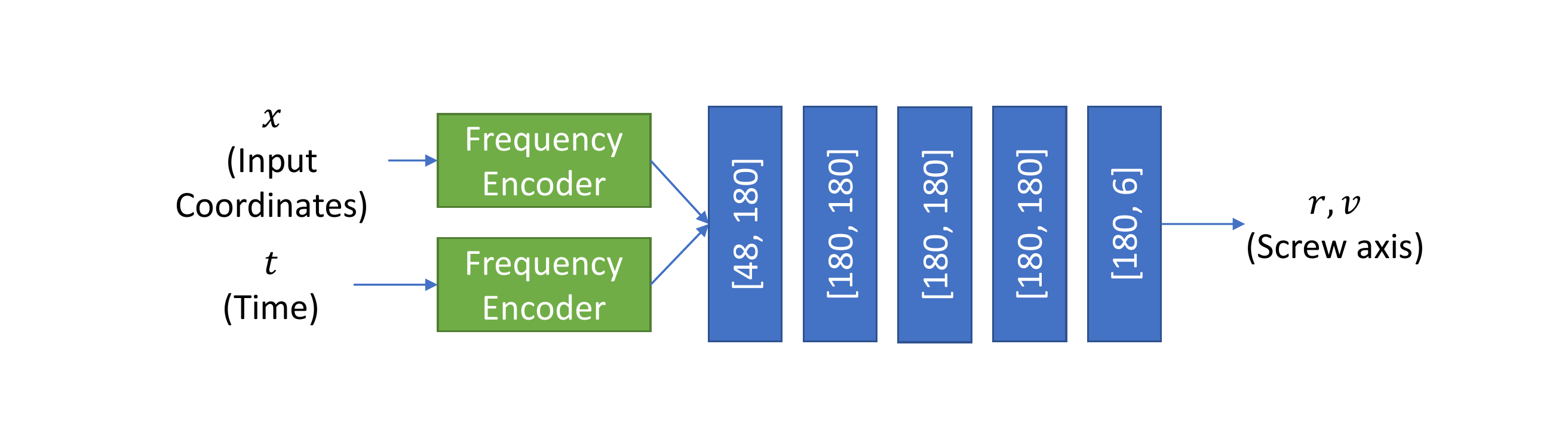}
    \caption{\SSDNerf deformation field: at each point $\point$ at time $t$, our deformation field outputs screw axis parameters $r, v$.}
    \label{fig:arch_deformation}
\end{figure}

Our implementation is based on~\cite{torch-ngp}, a re-implementation of~\cite{mueller2022instant}. We extend the NeRF architecture by having $M$ color and density outputs, where $M$ is the number of segmentation layers. Additionally, we extend it with a deformation field similar to by~\cite{park2021nerfies}, which uses a screw axis representation of rigid transformations. Similar to~\cite{park2021nerfies}, we use a regularizer on the Laplacian of the deformation field. \Figure{arch_new} shows our network architecture, and~\Figure{arch_deformation}. When $M=1$  (using a single class upon removing the semantic loss and the proposed regularizers), our network architecture %
reduces to~\cite{torch-ngp}.

\section{Derivation of Compositing Equations}\label{sec:compositing_derivation}
\subsection{NeRF \cite{mildenhall2020nerf} Image Reconstruction.} The compositing formula proposed in the original NeRF publication~\cite{mildenhall2020nerf} can be derived from the volume rendering integral~\cite{max1995volume}. We parameterize a ray as $\ray(t) = \mathbf{o} + t\mathbf{d}$, where $\mathbf{o}$ is the ray origin and $\mathbf{d}$ the ray direction. The expected color $C(\ray)$ with near and far bounds $t_n$ and $t_f$ is
\begin{align}
    C(\ray) &= \int_{t_n}^{t_f} T(t)\density(t)\radiance(t)dt, \nonumber\\ 
    & \text{where} \quad T(t) = \exp\left(-\int_{t_n}^t\density(s)ds\right).
\end{align}
We discretize the ray in $N$ segments and assume the color and density to be constant along the segment leading to 
\begin{align}
    C&(\ray) = \sum_{j=1}^N \int_{t_j}^{t_{j+1}}T(t) \density_j\radiance_j  dt \\
            &= \sum_{j=1}^N T(t_j) \left(\int_{t_j}^{t_{j+1}}\exp\left(-\int_{t_j}^t\density_jds\right) dt\right) \density_j\radiance_j  \\
            &= \sum_{j=1}^N T(t_j)  \left(\int_{t_j}^{t_{j+1}}\exp\left(-\density_j(t - t_j)\right)  dt\right) \density_j\radiance_j\\
            &= \sum_{j=1}^N T_j \left(1 - \exp\left(-\density_j\delta_j\right)\right)\radiance_j 
\end{align}
where $\delta_j = t_{j+1} - t_j $ is the length of the $j^\text{th}$ segment and $T_j = T(t_j) = \exp\left(-\sum_{k=1}^{j-1}\density_k\delta_k\right)$.

\subsection{\SSDNerf Image Reconstruction.} To decompose a NeRF into a set of $M$ semantic layers, our approach extends this formulation and generates a color and a density value per layer $F : (\point, \direction) \rightarrow (\radiance^1, \density^1, \hdots, \radiance^{M}, \density^{M})$. The compositing equation for $M$ layers then becomes 
\begin{align}
    C(\ray) &= \int_{t_n}^{t_f} T(t)\sum_{i=1}^{M}\density^i(t)\radiance^i(t)dt, \nonumber\\ 
    &\text{where} \quad T(t) = \exp\left(-\int_{t_n}^t\sum_{i=1}^{M}\density^i(s)ds\right).
\end{align}
Similarly, we discretize the ray into $N$ segments and assume the color and density to be constant along the segment, leading to 
\begin{align}
    &C(\ray) = \sum_{j=1}^N \int_{t_j}^{t_{j+1}}T(t) \sum_{i=1}^{M}\density^i_j\radiance^i_j  dt \\
            =& \sum_{j=1}^N T(t_j) \left(\int_{t_j}^{t_{j+1}}\exp\left(-\int_{t_j}^t\sum_{i=1}^{M}\density^i_jds\right)dt\right)\sum_{i=1}^{M}\density^i_j\radiance^i_j\\
            =& \sum_{j=1}^N T(t_j) \left(\int_{t_j}^{t_{j+1}}\exp\left(-\sum_{i=1}^{M}\density^i_j(t - t_j) \right) dt\right) \sum_{i=1}^{M}\density^i_j\radiance^i_j\\
            =& \sum_{j=1}^N T_j \left(1 - \exp\left(-\sum_{i=1}^{M}\density^i_j\delta_j\right)\right)\frac{\sum_{i=1}^{M}\density^i_j\radiance^i_j }{\sum_{i=1}^{M}\density^i_j} ,
\end{align}
where $T_j = T(t_j) = \exp\left(-\sum_{k=1}^{j-1}\sum_{i=1}^{M}\density^i_k\delta_k\right)$.

\subsection{\SSDNerf Layer Reconstruction.} Generating the $i^\text{th}$ layer can be easily achieved by keeping the density of the $i^\text{th}$ layer the same and setting the density of other layers to $0$ leading to 
\begin{align}
    C^i(\ray) &= \sum_{j=1}^N T^i_j \left(1 - \exp\left(-\density^i_j\delta_j\right)\right)\radiance^i_j,
\end{align}
where $T^i_j = \exp\left(-\sum_{k=1}^{j-1}\density^i_k\delta_k\right)$. In a similar spirit, generating the $i^\text{th}$ segmentation mask can be achieved by setting the color to $1$ for the $i^\text{th}$ layer and the color of the other layers to $0$ leading to 
\begin{align}
    S^i(\ray) &= \sum_{j=1}^N T_j \left(1 - \exp\left(-\sum_{i=1}^{M}\density^i_j\delta_j\right)\right)\frac{\density^i_j}{\sum_{i=1}^{M}\density^i_j}.
\end{align}

\subsection{SNeRF Layer Reconstruction.}\label{subsec:snerf_layers_supp} The original implementation of SNeRF~\cite{Zhi21semanticnerf} does not generate disentangled segmentation layers. To compare our approach to SNeRF we modify its compositing equation to remove the radiance contribution of some 3D points based on the predicted semantic class probabilities at those points. More formally, we reweight the density by the class probability leading to 
\begin{align}
    C^i(\ray) &= \int_{t_n}^{t_f} T^i(t)p^i(t)\density(t)\radiance(t)dt, \nonumber\\
    &\text{where} \quad T^i(t) = \exp\left(-\int_{t_n}^tp^i(s)\density(s)ds\right),
\end{align}
where $p^i(t)$ is the probability of the $i^\text{th}$ class at location $t$. We discretize the ray into $N$ segments and assume the color, density, and probability to be constant along the segment, leading to 
\begin{align}
    &C^i(\ray) = \sum_{j=1}^N \int_{t_j}^{t_{j+1}}T^i(t) p^i_j\density_j\radiance_j  dt \\
        =& \sum_{j=1}^N T^i(t_j) \left(\int_{t_j}^{t_{j+1}}\exp\left(-\int_{t_j}^tp^i_j\density_jds\right)dt\right)p^i_j\density_j\radiance_j\\
        =& \sum_{j=1}^N T^i(t_j) \left(\int_{t_j}^{t_{j+1}}\exp\left(-p^i_j\density_j(t - t_j) \right) dt\right) p^i_j\density_j\radiance_j\\
        =& \sum_{j=1}^N T^i_j \left(1 - \exp\left(-p^i_j\density_j\delta_j\right)\right)\radiance_j,
\end{align}
where $T^i_j = T^i(t_j) = \exp\left(-\sum_{k=1}^{j-1}p^i_k\density_k\delta_k\right)$.

{\small
\bibliographystyle{ieee_fullname}
\bibliography{egbib}

\begin{thebibliography}{10}\itemsep=-1pt

\bibitem{Aksoy2018sss}
Ya\u{g}{\i}z Aksoy, Tae-Hyun Oh, Sylvain Paris, Marc Pollefeys, and Wojciech
  Matusik.
\newblock Semantic soft segmentation.
\newblock {\em TOG}, 2018.

\bibitem{torf}
Benjamin Attal, Eliot Laidlaw, Aaron Gokaslan, Changil Kim, Christian Richardt,
  James Tompkin, and Matthew O'Toole.
\newblock T{\"o}rf: Time-of-flight radiance fields for dynamic scene view
  synthesis.
\newblock {\em NeurIPS}, 2021.

\bibitem{barron2021mipnerf360}
Jonathan~T. Barron, Ben Mildenhall, Dor Verbin, Pratul~P. Srinivasan, and Peter
  Hedman.
\newblock {Mip-NeRF 360}: Unbounded anti-aliased neural radiance fields.
\newblock {\em arXiv}, 2021.

\bibitem{chen2017deeplab}
Liang-Chieh Chen, George Papandreou, Iasonas Kokkinos, Kevin Murphy, and
  Alan~L. Yuille.
\newblock {DeepLab}: Semantic image segmentation with deep convolutional nets,
  atrous convolution, and fully connected {CRFs}.
\newblock {\em TPAMI}, 2017.

\bibitem{dellaert2020neural}
Frank Dellaert and Lin Yen-Chen.
\newblock Neural volume rendering: Nerf and beyond.
\newblock {\em arXiv}, 2020.

\bibitem{elich2020semsup3dobjs}
Cathrin Elich, Martin~R Oswald, Marc Pollefeys, and Joerg Stueckler.
\newblock Weakly supervised learning of multi-object 3d scene decompositions
  using deep shape priors.
\newblock {\em CoRR}, 2020.

\bibitem{nerface}
Guy Gafni, Justus Thies, Michael Zollh{\"o}fer, and Matthias Nie{\ss}ner.
\newblock Dynamic neural radiance fields for monocular {4D} facial avatar
  reconstruction.
\newblock In {\em CVPR}, 2021.

\bibitem{guler2018densepose}
R{\i}za~Alp G{\"u}ler, Natalia Neverova, and Iasonas Kokkinos.
\newblock Densepose: Dense human pose estimation in the wild.
\newblock In {\em CVPR}, 2018.

\bibitem{guo2020object}
Michelle Guo, Alireza Fathi, Jiajun Wu, and Thomas Funkhouser.
\newblock Object-centric neural scene rendering.
\newblock {\em arXiv}, 2020.

\bibitem{adnerf}
Yudong Guo, Keyu Chen, Sen Liang, Yongjin Liu, Hujun Bao, and Juyong Zhang.
\newblock {AD-NeRF}: Audio driven neural radiance fields for talking head
  synthesis.
\newblock In {\em ICCV}, 2021.

\bibitem{hackel2016fast}
Timo Hackel, Jan~D Wegner, and Konrad Schindler.
\newblock Fast semantic segmentation of {3D} point clouds with strongly varying
  density.
\newblock {\em ISPRS annals of the photogrammetry, remote sensing and spatial
  information sciences}, 2016.

\bibitem{hane2016dense}
Christian H{\"a}ne, Christopher Zach, Andrea Cohen, and Marc Pollefeys.
\newblock Dense semantic {3D} reconstruction.
\newblock {\em TPAMI}, 2016.

\bibitem{he2017mask}
Kaiming He, Georgia Gkioxari, Piotr Doll{\'a}r, and Ross Girshick.
\newblock Mask {R-CNN}.
\newblock In {\em ICCV}, 2017.

\bibitem{hou20193d}
Ji Hou, Angela Dai, and Matthias Nie{\ss}ner.
\newblock {3D-SIS}: {3D} semantic instance segmentation of {RGB-D} scans.
\newblock In {\em CVPR}, 2019.

\bibitem{isack2020repose}
Hossam Isack, Christian Haene, Cem Keskin, Sofien Bouaziz, Yuri Boykov, Shahram
  Izadi, and Sameh Khamis.
\newblock Repose: Learning deep kinematic priors for fast human pose
  estimation.
\newblock {\em arXiv}, 2020.

\bibitem{raytracing}
James~T. Kajiya and Brian~P Von~Herzen.
\newblock Ray tracing volume densities.
\newblock In {\em SIGGRAPH}, 1984.

\bibitem{KingmaB14}
Diederik~P. Kingma and Jimmy Ba.
\newblock Adam: {A} method for stochastic optimization.
\newblock In Yoshua Bengio and Yann LeCun, editors, {\em ICLR}, 2015.

\bibitem{Levin08matting}
Anat Levin, Dani Lischinski, and Yair Weiss.
\newblock A closed-form solution to natural image matting.
\newblock {\em TPAMI}, 2008.

\bibitem{Levin08spectralmatting}
Anat Levin, Alex Rav{-}Acha, and Dani Lischinski.
\newblock Spectral matting.
\newblock {\em TPAMI}, 2008.

\bibitem{Lombardi21}
Stephen Lombardi, Tomas Simon, Gabriel Schwartz, Michael Zollhoefer, Yaser
  Sheikh, and Jason Saragih.
\newblock Mixture of volumetric primitives for efficient neural rendering.
\newblock {\em TOG}, 2021.

\bibitem{max1995volume}
N. Max.
\newblock Optical models for direct volume rendering.
\newblock {\em TVCG}, 1995.

\bibitem{mildenhall2020nerf}
Ben Mildenhall, Pratul~P. Srinivasan, Matthew Tancik, Jonathan~T. Barron, Ravi
  Ramamoorthi, and Ren Ng.
\newblock {NeRF}: Representing scenes as neural radiance fields for view
  synthesis.
\newblock In {\em ECCV}, 2020.

\bibitem{mueller2022instant}
Thomas M\"uller, Alex Evans, Christoph Schied, and Alexander Keller.
\newblock Instant neural graphics primitives with a multiresolution hash
  encoding.
\newblock {\em ACM Trans. Graph.}, 41(4):102:1--102:15, July 2022.

\bibitem{niemeyer2021giraffe}
Michael Niemeyer and Andreas Geiger.
\newblock Giraffe: Representing scenes as compositional generative neural
  feature fields.
\newblock In {\em CVPR}, 2021.

\bibitem{narf}
Atsuhiro Noguchi, Xiao Sun, Stephen Lin, and Tatsuya Harada.
\newblock Neural articulated radiance field.
\newblock In {\em ICCV}, 2021.

\bibitem{park2021nerfies}
Keunhong Park, Utkarsh Sinha, Jonathan~T. Barron, Sofien Bouaziz, Dan~B
  Goldman, Steven~M. Seitz, and Ricardo Martin-Brualla.
\newblock Nerfies: Deformable neural radiance fields.
\newblock {\em ICCV}, 2021.

\bibitem{animatablenerf}
Sida Peng, Junting Dong, Qianqian Wang, Shangzhan Zhang, Qing Shuai, Xiaowei
  Zhou, and Hujun Bao.
\newblock Animatable neural radiance fields for modeling dynamic human bodies.
\newblock In {\em ICCV}, 2021.

\bibitem{neuralbody}
Sida Peng, Yuanqing Zhang, Yinghao Xu, Qianqian Wang, Qing Shuai, Hujun Bao,
  and Xiaowei Zhou.
\newblock Neural body: Implicit neural representations with structured latent
  codes for novel view synthesis of dynamic humans.
\newblock In {\em CVPR}, 2021.

\bibitem{dnerf}
Albert Pumarola, Enric Corona, Gerard Pons-Moll, and Francesc Moreno-Noguer.
\newblock {D-NeRF}: Neural radiance fields for dynamic scenes.
\newblock {\em arXiv}, 2020.

\bibitem{rebain2020derf}
Daniel Rebain, Wei Jiang, Soroosh Yazdani, Ke Li, Kwang~Moo Yi, and Andrea
  Tagliasacchi.
\newblock {DeRF}: Decomposed radiance fields.
\newblock In {\em CVPR}, 2020.

\bibitem{reizenstein21co3d}
Jeremy Reizenstein, Roman Shapovalov, Philipp Henzler, Luca Sbordone, Patrick
  Labatut, and David Novotny.
\newblock Common objects in 3d: Large-scale learning and evaluation of
  real-life 3d category reconstruction.
\newblock In {\em International Conference on Computer Vision}, 2021.

\bibitem{yu_and_fridovichkeil2021plenoxels}
{Sara Fridovich-Keil and Alex Yu}, Matthew Tancik, Qinhong Chen, Benjamin
  Recht, and Angjoo Kanazawa.
\newblock Plenoxels: Radiance fields without neural networks.
\newblock In {\em CVPR}, 2022.

\bibitem{savinov2016semantic}
Nikolay Savinov, Christian Hane, Lubor Ladicky, and Marc Pollefeys.
\newblock Semantic {3D} reconstruction with continuous regularization and ray
  potentials using a visibility consistency constraint.
\newblock In {\em CVPR}, 2016.

\bibitem{Sbai20unsuperdecomp}
Othman Sbai, Camille Couprie, and Mathieu Aubry.
\newblock Unsupervised image decomposition in vector layers.
\newblock In {\em ICIP}, 2020.

\bibitem{schoenberger2016sfm}
Johannes~Lutz Sch\"{o}nberger and Jan-Michael Frahm.
\newblock Structure-from-motion revisited.
\newblock In {\em Conference on Computer Vision and Pattern Recognition
  (CVPR)}, 2016.

\bibitem{sengupta2013urban}
Sunando Sengupta, Eric Greveson, Ali Shahrokni, and Philip~HS Torr.
\newblock Urban {3D} semantic modelling using stereo vision.
\newblock In {\em ICRA}, 2013.

\bibitem{sridhar2019multiview}
Srinath Sridhar, Davis Rempe, Julien Valentin, Bouaziz Sofien, and Leonidas~J
  Guibas.
\newblock Multiview aggregation for learning category-specific shape
  reconstruction.
\newblock {\em NeurIPS}, 2019.

\bibitem{stelzner2021decomposing}
Karl Stelzner, Kristian Kersting, and Adam~R Kosiorek.
\newblock Decomposing {3D} scenes into objects via unsupervised volume
  segmentation.
\newblock {\em arXiv}, 2021.

\bibitem{sun2021direct}
Cheng Sun, Min Sun, and Hwann-Tzong Chen.
\newblock Direct voxel grid optimization: Super-fast convergence for radiance
  fields reconstruction.
\newblock {\em arXiv}, 2021.

\bibitem{SzeliskiBook}
Richard Szeliski.
\newblock {\em Computer Vision - Algorithms and Applications, Second Edition}.
\newblock Springer, 2022.

\bibitem{torch-ngp}
Jiaxiang Tang.
\newblock Torch-ngp: a pytorch implementation of instant-ngp, 2022.
\newblock https://github.com/ashawkey/torch-ngp.

\bibitem{tianStrikingRightBalance2022}
Junjiao Tian, Niluthpol~Chowdhury Mithun, Zachary Seymour, Han-Pang Chiu, and
  Zsolt Kira.
\newblock Striking the {{Right Balance}}: {{Recall Loss}} for {{Semantic
  Segmentation}}.
\newblock In {\em 2022 {{International Conference}} on {{Robotics}} and
  {{Automation}} ({{ICRA}})}, pages 5063--5069, 2022.

\bibitem{tretschk2021nonrigid}
Edgar Tretschk, Ayush Tewari, Vladislav Golyanik, Michael Zollh\"{o}fer,
  Christoph Lassner, and Christian Theobalt.
\newblock Non-rigid neural radiance fields: Reconstruction and novel view
  synthesis of a dynamic scene from monocular video.
\newblock In {\em ICCV}, 2021.

\bibitem{tulsiani2017multi}
Shubham Tulsiani, Tinghui Zhou, Alexei~A Efros, and Jitendra Malik.
\newblock Multi-view supervision for single-view reconstruction via
  differentiable ray consistency.
\newblock In {\em CVPR}, 2017.

\bibitem{valentin2015semanticpaint}
Julien Valentin, Vibhav Vineet, Ming-Ming Cheng, David Kim, Jamie Shotton,
  Pushmeet Kohli, Matthias Nie{\ss}ner, Antonio Criminisi, Shahram Izadi, and
  Philip Torr.
\newblock {SemanticPaint}: Interactive {3D} labeling and learning at your
  fingertips.
\newblock {\em TOG}, 2015.

\bibitem{wang2021head}
Ziyan Wang, Timur Bagautdinov, Stephen Lombardi, Tomas Simon, Jason Saragih,
  Jessica Hodgins, and Michael Zollhofer.
\newblock Learning compositional radiance fields of dynamic human heads.
\newblock In {\em CVPR}, 2021.

\bibitem{xiangli2021citynerf}
Yuanbo Xiangli, Linning Xu, Xingang Pan, Nanxuan Zhao, Anyi Rao, Christian
  Theobalt, Bo Dai, and Dahua Lin.
\newblock {CityNeRF}: Building {NeRF} at city scale.
\newblock {\em arXiv}, 2021.

\bibitem{yu2021unsupervised}
Hong-Xing Yu, Leonidas~J Guibas, and Jiajun Wu.
\newblock Unsupervised discovery of object radiance fields.
\newblock In {\em ICLR}, 2022.

\bibitem{nerf++}
Kai Zhang, Gernot Riegler, Noah Snavely, and Vladlen Koltun.
\newblock {NeRF++}: Analyzing and improving neural radiance fields.
\newblock {\em arXiv}, 2020.

\bibitem{Zhi21semanticnerf}
Shuaifeng Zhi, Tristan Laidlow, Stefan Leutenegger, and Andrew Davison.
\newblock In-place scene labelling and understanding with implicit scene
  representation.
\newblock In {\em ICCV}, 2021.

\end{thebibliography}
}

\end{document}